\documentclass[10pt,twocolumn,letterpaper]{article}
\usepackage{cvpr}
\usepackage{times}
\usepackage{epsfig}
\usepackage{graphicx}
\usepackage{amsmath}
\usepackage{amssymb}
\usepackage{subfigure}
\usepackage{algorithmic,algorithm}
\usepackage{booktabs}

\newcommand{\1}{\mbox{1}\hspace{-0.25em}\mbox{l}}
\newtheorem{theorem1}{Theorem}
\newcommand{\hdistance}{\mathcal{H}\Delta\mathcal{H}}

\usepackage[pagebackref=true,breaklinks=true,letterpaper=true,colorlinks,bookmarks=false]{hyperref}

\cvprfinalcopy % *** Uncomment this line for the final submission

\ifcvprfinal\fi
\begin{document}

%%%%%%%%% TITLE
\title{TWINs: Two Weighted Inconsistency-reduced\\Networks for Partial Domain Adaptation}
\author{Toshihiko Matsuura\thanks{Authors contributed equally.}\ \,$^1$ Kuniaki Saito$^\ast$$^2$ Tatsuya Harada$^{1,3}$ \\$^1$The University of Tokyo, $^2$Boston University, $^3$RIKEN\\
\tt\small $^1$\{matsuura, harada\}@mi.t.u-tokyo.ac.jp, $^2$\{keisaito\}@bu.edu}
% \author{Toshihiko Matsuura\\
% The University of Tokyo\\
% {\tt\small matsuura@mi.t.u-tokyo.ac.jp}
% \and
% Kuniaki Saito \\
% Boston University\\
% {\tt\small keisaito@bu.edu}
% \and 
% Tatsuya Harada \\
% The University of Tokyo, RIKEN\\
% {\tt \small harada@mi.t.u-tokyo.ac.jp}
% }

\maketitle
%\thispagestyle{empty}

%%%%%%%%% ABSTRACT
\begin{abstract}
The task of unsupervised domain adaptation is proposed to transfer the knowledge of a label-rich domain (source domain) to a label-scarce domain (target domain). 
Matching feature distributions between different domains is a widely applied method for the aforementioned task. However, the method does not perform well when classes in the two domains are not identical. Specifically, when the classes of the target correspond to a subset of those of the source, target samples can be incorrectly aligned with the classes that exist only in the source. This problem setting is termed as partial domain adaptation (PDA). 
In this study, we propose a novel method called Two Weighted Inconsistency-reduced Networks (TWINs) for PDA. We utilize two classification networks to estimate the ratio of the target samples in each class with which a classification loss is weighted to adapt the classes present in the target domain. Furthermore, to extract discriminative features for the target, we propose to minimize the divergence between domains measured by the classifiers' inconsistency on target samples. We empirically demonstrate that reducing the inconsistency between two networks is effective for PDA and that our method outperforms other existing methods with a large margin in several datasets.
\end{abstract}

%%%%%%%%% BODY TEXT
\begin{figure}
\begin{center}
  \includegraphics[clip,width=0.98\linewidth]{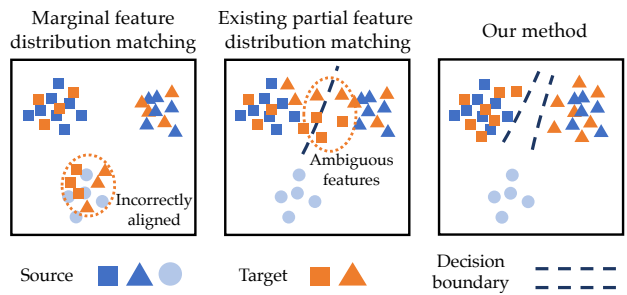}
\end{center}
  \caption{Comparison of our method to %marginal feature distribution matching methods and existing partial matching methods. 
  existing methods. Left: marginal feature distribution matching methods incorrectly align target features with the class of the source that is absent in the target domain. Center: Existing partial feature distribution matching methods do not consider the task-specific classifier's boundary for adaptation that tends to generate ambiguous features. Right: Our method considers the decision boundary to align target features with the source, and this leads to the extraction of discriminative features.}
\label{fig:feature}
\end{figure}
\section{Introduction}
Deep convolutional neural networks trained on a large number of labeled data boost the performance of image recognition on various tasks. However, the preparation of many labeled samples to train the network is time-consuming and expensive. The method for transferring knowledge from a label-rich domain (source domain) to a label-scarce domain (target domain) is termed as domain adaptation and enables us to reduce the cost for annotation.  Specifically, the method for unsupervised domain adaptation (UDA) where we do not require any annotated target samples during training can solve the aforementioned difficulty.
The difficulty in the task involves the difference between each domain with respect to the texture, color, and appearance of objects. The classifier trained on the source domain typically does not work well on the other domain due to the domain-gap problem. Additionally, the target domain includes only unlabeled samples, and this implies that we do not know the classes that are present in the target domain.
% Subsequently, a possible strategy involves collecting samples belonging to various classes subsuming those present in the target domain and adapting a model from the source domain to the target domain that may not include a few some classes present in the source domain.
A possible strategy involves collecting samples belonging to various classes subsuming those present in the target domain and adapting a model from the source to the target domain.
As the result, the target domain may not include some classes present in the source domain. The adaptation setting is termed as \textit{partial domain adaptation} (PDA) and corresponds to an extremely practical setting.\par
However, several methods for UDA~\cite{GRL,DAN,MCD,domain_confusion} should degrade their performance in the PDA setting because they assume that the source and target completely share the same classes.
Their aim involves matching the marginal feature distributions between different domains. If the target feature distribution is aligned with the source overall, the target samples can be assigned to the class absent in the target domain as shown in the left of Fig.~\ref{fig:feature}. In summary, partially aligning marginal feature distributions is necessary in PDA.\par
Our method integrates three motivations to effectively deal with PDA. The first motivation involves precisely estimating the label distribution of the target to train a model on the classes present in the target domain.
Second, the extraction of discriminative features for the target domain is important for highly accurate classification of target samples as shown in the center and right of Fig.~\ref{fig:feature}. This figure indicates that considering relationship between target samples and task-specific decision boundaries is important to extract discriminative features. 
The utilization of a classifier's output for the target domain is shown to be effective to extract discriminative features~\cite{MCD} although the method matches marginal feature distributions. 
Third, to partially match feature distributions between domains, it is necessary to use a measurement that can partially evaluate the distance between domains. Many methods for UDA aim to measure the distance between the entire distributions of the source and target, which is not desirable in PDA.\par 
In this paper, we propose a novel method called \textit{Two Weighted Inconsistency-reduced Networks} (TWINs). We utilize two classification networks that do not share their parameters. With respect to the first motivation, the label distribution of the target is precisely estimated by the outputs of the two networks for all target samples with which a classification loss is weighted. The estimation is more accurate than when using one network. With respect to the second and third motivations, we propose to minimize the domain divergence measured by the inconsistency of classifiers on target samples. Thus, the two networks are trained to agree on their predictions on the target samples to extract discriminative features. To partially measure the distance between domains, we propose not to use adversarial training, and this is different from \cite{MCD}. 
Our method displays a connection with the theory of domain adaptation that measures the divergence between domains by the inconsistency of two classifiers.\par
We evaluate our method on digits, traffic signs, and object classification tasks on PDA setting. In most tasks, our method outperforms other existing methods by a large margin.
%-------------------------------------------------------------------------

\section{Related Work}
\subsection{Unsupervised Domain Adaptation}
Several methods are proposed for unsupervised domain adaptation (UDA). In the present study, we mainly focus on methods that are based on deep learning, since they are proven as powerful learning systems.\par
\textbf{Feature distribution matching} to extract domain invariant features is the most popular method for UDA and includes Maximum Mean Discrepancy~\cite{MMD,DAN,JAN,RTN,WMMD}, Central Moment Discrepancy~\cite{CMD}, and CORAL~\cite{CORAL}.\par
\textbf{Domain classifier based method} through adversarial learning is also a representative method for UDA~\cite{GRL,CADA,LEL,CoGAN,MCD,domain_confusion,ADDA}. A domain classifier is trained to discriminate the domain from where the feature originates while a feature extractor is trained to deceive the domain classifier. Adversarial training aligns the feature distribution of the source and target with each other.
The methods are designed with the assumption that the label distributions of the source and target are approximately the same. When the assumption does not hold, such as in the case of partial domain adaptation, the target samples can be assigned to the class of the source that is absent in the target domain.\par
\textbf{Classifier's discrepancy based method} is recently proposed for UDA and significantly improves performance. Maximum Classifier Discrepancy~\cite{MCD} utilizes the outputs of task-specific classifiers to align features and also to extract more discriminative features for target samples. They construct two classifier networks with a shared feature extractor network. They train the two classifiers to output different predictions on the target samples while training the feature extractor to generate features that make the output of the two networks similar.
The method is useful since it considers the task-specific decision boundary and avoids generating ambiguous target features near class boundaries. Sampling two classifiers by using dropout achieves a similar effect~\cite{dropout}. However, the methods are not also effective for partial domain adaptation (PDA). They use adversarial training between classifiers and feature extractor, with which source and target features are likely to be strictly aligned. We discuss further details of the aforementioned point in Sec.~\ref{sec:method}.\par
In the study, we introduce a method that can partially adapt features by weighting the classification loss by the estimated label distribution of the target domain and training to reduce inconsistency between two task-specific classifiers. Please note that our method does not rely on adversarial training. Deep mutual learning is proposed for large-scale supervised image classification~\cite{dml}. This method also has the objective of minimizing inconsistency of two networks. They do not show that the technique is useful for UDA. In addition, in our work, we present how to utilize the technique for PDA and why it is useful for this task.\par

\begin{figure*}
\centering
\includegraphics[clip,width=0.9\linewidth]{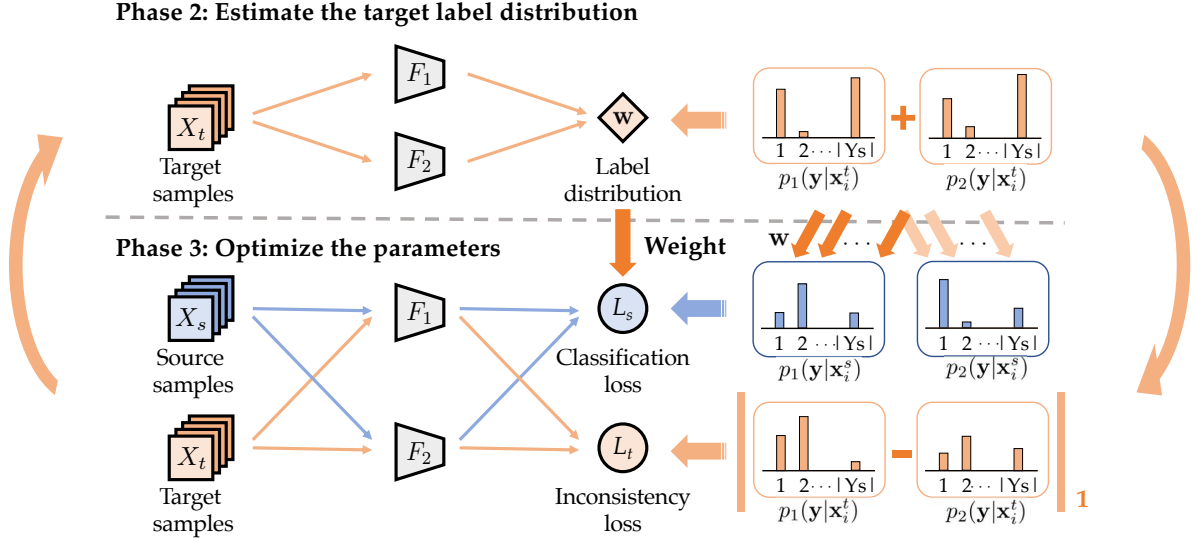}
\caption{The overview of the proposed method where $X_s$ and $X_t$ denote source and target samples, respectively; and $F_1$ and $F_2$ denote two classifiers. The label distribution of the target domain ${\mathbf w}$ is estimated by two classifiers' outputs on target samples. The estimated label distribution ${\mathbf w}$ is used to weight classification loss on source samples ($L_s$). Inconsistency loss $L_t$ on target samples are also calculated by the difference between two classifiers' outputs.}
\label{fig:training}
\end{figure*}

\subsection{Partial Domain Adaptation}
In the PDA setting, the target domain contains the classes that are a subset of the source classes. To the best of our knowledge, all methods for PDA utilize a domain classifier to achieve adaptation~\cite{SAN,PADA,ImportanceWA}. The main idea of the aforementioned methods is to identify whether source samples belong to the classes present in the target domain and weight the task-specific classifier's loss or the domain classifier's loss to avoid the alignment of target samples with the source classes absent in the target domain. %Importance Weighted Adversarial Nets (IWAN)~\cite{ImportanceWA} trains two domain classifiers wherein one domain classifier plays an auxiliary role to create weights for the other domain classifier. %and to create weights for the other domain classifier. The re-weighting allows the alignment of target samples with only source samples similar to the target. 
For instance, Partial Adversarial Domain Adaptation (PADA)~\cite{PADA} estimates the label distribution of the target samples by averaging the outputs of the classifier for target samples and utilizes it to re-weight the task-specific classification loss and domain classification loss.
The major differences in the methods are as follows. First, we estimate the label distribution of the target samples by two parameter-unshared networks that enables accurate estimation. Second, we introduce the idea of a task-specific classifier based on feature distribution alignment for the task. Existing methods do not use task-specific classifiers to align features that generate ambiguous features near the decision boundary.

\section{Proposed Method}\label{sec:method}
This section presents the proposed method for partial domain adaptation in detail. The overview of our method is illustrated in Fig.~\ref{fig:training}. The following two key ideas are involved in our method: classification loss on the source weighted by the estimated label distribution in the target domain and feature distribution alignment by using task-specific classifiers. Additionally, we propose a training procedure to integrate the two ideas to achieve effective adaptation.
The label distribution of the target domain is estimated by using two networks that do not share parameters. When the estimated label distribution is obtained, it is used to make the networks focus on classifying classes present in the target domain as described in Sec.~\ref{sec:weighted_loss}. Specifically, the distribution is used to weight the classification loss on the source samples.
Furthermore, we conduct feature distribution alignment by minimizing the inconsistency of the two task-specific classifiers, and this leads to the extraction of discriminative features (and not ambiguous features) for target samples as described in Sec.~\ref{sec:inconsistency_loss}.
The estimation of the label distribution and feature distribution alignment are alternately performed. All the training procedures are described in Sec.~\ref{sec:procedure}.
We make a connection between our method and the theory of domain adaptation in Sec.~\ref{sec:theory}.\par
We state the definitions of terminologies. The source domain data $X_s \in \mathbb{R}^{d\times n_s}$ are drawn from distribution $P_s\left(X_s\right)$, and target domain data $X_t \in \mathbb{R}^{d\times n_t}$ are drawn from distribution $P_t\left(X_t\right)$ where $d$ denotes the dimension of the data instance, and $n_s$ and $n_t$ denote the number of samples in the source and target domain respectively. This is due to the domain shift, $P_s\left(X_s\right) \neq P_t\left(X_t\right)$. We use labeled source samples $\mathcal{D}_s=\left\{\left({\mathbf{x}}_i^s,y_i^s\right)\right\}_{i=1}^{n_s}, {\mathbf{x}}_i^s \in \mathbb{R}^d$ and unlabeled target samples $\mathcal{D}_t=\left\{\left({\mathbf{x}}_i^t\right)\right\}_{i=1}^{n_t}, {\mathbf{x}}_i^t \in \mathbb{R}^d$ during training. The source label space $\mathcal{Y}_s$ and the target one $\mathcal{Y}_t$ are different. The target domain label space is contained in the source domain label space ($\mathcal{Y}_t \subseteq \mathcal{Y}_s$). We use two networks, namely $F_1$ and $F_2$, with exactly the same architecture although they do not share their parameters $\theta_1,\theta_2$.
The probabilities that ${\mathbf x}$ is classified into class $k$ when it is inputted into $F_1$ and $F_2$ are denoted by $p_1(y=k|{\mathbf x})$ and $p_2(y=k|{\mathbf x})$, respectively. Furthermore, we use the notation $p_1({\mathbf y}|{\mathbf x})$ and $p_2({\mathbf y}|{\mathbf x})$ to denote the $|\mathcal{Y}_s|$-dimensional probabilistic output for ${\mathbf x}$ inputted into $F_1$ and $F_2$, respectively. We assume that the outputs are obtained after the softmax layer.
\par

\subsection{Weighted Loss with the Label Distribution}\label{sec:weighted_loss}
\noindent
\textbf{Target Label-Distribution Estimation.}\ \ We explain how to estimate the label distribution of the target domain.
As we mentioned, we assume that there are two networks that are trained with the following classification loss on the source domain:
\begin{align}
L_s(X_s,Y_s)\mathalpha{=}L_{s_1}\left(F_1\left(X_s\right),Y_s\right)
\mathalpha{+}L_{s_2}\left(F_2\left(X_s\right),Y_s\right),\label{eq:loss}
\end{align}
where $L_{s_1}$ and $L_{s_2}$ denote the classification losses with respect to $F_1$ and $F_2$, respectively.
$L_{s_j}(j=1,2)$ is defined as follows, 
\begin{equation}
    L_{s_j}=-\frac{1}{n_s}\sum_{i=1}^{n_s}\sum_{k=1}^{|\mathcal{Y}_s|}\1_{[k=y_i^s]}\log p_j\left(y=k|{\mathbf x}_i^s\right),
    \label{eq:entropy}
\end{equation}
where $\1_{[k=y_s]}$ is $1$ when $k=y_s$, otherwise, $0$.\par
We focus on the estimation of the label distribution by averaging the outputs of two networks for all target samples. Specifically, ${\mathbf w}$ is calculated as, 
\begin{equation}
\mathbf{w}=\frac{|\mathcal{Y}_s|}{2n_t}\sum_{i=1}^{n_t}\left(p_1(\mathbf{y}|\mathbf{x}_i^t)+p_2(\mathbf{y}|\mathbf{x}_i^t)\right),\label{eq:weight}
\end{equation}
where ${\mathbf w}$ denotes a $|\mathcal{Y}_s|$-dimensional vector. To obtain the weight, we multiply the label distribution by the number of the source class to prevent the loss from being very small.\par
\noindent
\textbf{Target Label-Distribution Weighted Loss.}\ \ The estimated label distribution is used to make the model focus on classifying classes present in the target domain.
Each dimension of the vector indicates the approximated ratio of the target samples of the corresponding class. While training the networks with source samples, we aim to focus on the present classes and suppress the effect of absent classes in the target. 
The weighted classification loss $L_s'$ is summarized as follows, 

\begin{align}
    L_s^{\prime}&= L_{s_1}^{\prime}(F_1(X_s),Y_s)+L_{s_2}^{\prime}(F_2(X_s),Y_s),
    \label{eq:loss_weighted}\\
    L_{s_j}^{\prime}&=-\frac{1}{n_s}\sum_{i=1}^{n_s}\sum_{k=1}^{|\mathcal{Y}_s|}w_k\1_{[k=y_s]}\log p_j\left(y=k|{\mathbf x}_i^s\right),
    \label{eq:entropy_weighted}
\end{align}
where $w_k$ denotes the k-th element of $\mathbf{w}$. 
With the loss function, the classifiers can increasingly focus on classes present in the target domain when trained on source samples.

\begin{figure}
\centering
\includegraphics[clip,width=0.7\linewidth]{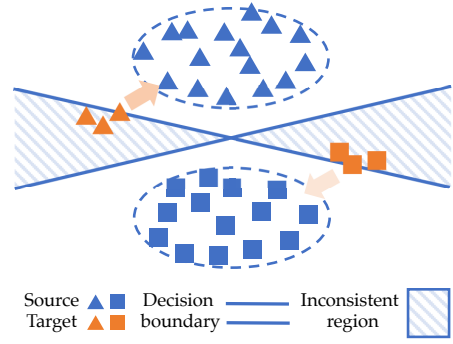}
\caption{The inconsistent region marked with diagonal lines denotes the area where the outputs of the two classifiers are inconsistent and the features should be far from the source samples. Reducing the inconsistency between the two classifiers' outputs aligns target with source samples considering task-specific decision boundaries.}
\label{fig:inconsistent}
\end{figure}

\subsection{Inconsistency Loss}\label{sec:inconsistency_loss}
We explain how we align the target samples with the source. We train $F_1$ and $F_2$ to reduce the inconsistency of predictions for target samples.
As shown in the study of MCD \cite{MCD} and Fig.~\ref{fig:inconsistent}, we can measure how discriminative the target features are by examining the inconsistency of the task-specific classifiers. If the inconsistency is high, the features should be far from the source. Conversely, if the inconsistency is low, the features should be near the source with respect to the task-specific decision boundary.\par
Then, we propose using an objective to minimize the inconsistency between the predictions of two classifiers for target samples and call it \textit{Inconsistency loss}. We make a connection with the theory of domain adaptation in Sec.~\ref{sec:theory}. In the study, we utilize the $L_1$ distance as inconsistency loss following \cite{MCD} although other functions can be used here. The inconsistency loss is as follows:
\begin{equation}
    L_t = \frac{1}{n_t} \sum_{i=1}^{n_t}\|p_1\left({\mathbf y}|{\mathbf x}_i^t\right)-p_2\left({\mathbf y}|{\mathbf x}_i^t\right)\|_1.
    \label{eq:loss_t}
\end{equation}

Our method is different from MCD~\cite{MCD} since the MCD includes a training step that increases the discrepancy of the task-specific classifiers. This is intended to effectively measure the distance between domains, and this should lead to strictly matching the feature distribution. Hence, the aforementioned type of strict matching should not be effective for PDA. This fact is empirically demonstrated in the experiments. Then, we propose not to employ the step of increasing the discrepancy of the task-specific classifiers.

\begin{algorithm}[t]
\caption{Training of TWINs. $N_1$, $N_2$, $N_3$ denote maximum iterations of Phase 1, the number of interval iterations of Phase 2, and maximum iterations of Phase 3, respectively.}
\label{alg:twins}
\begin{algorithmic}%[1]
\REQUIRE
Data: $\mathcal D_s=\left\{\left({\mathbf{x}}_i^s,y_i^s\right)\right\}_{i=1}^{n_s},\mathcal{D}_t=\left\{{\mathbf x}_i^t\right\}_{i=1}^{n_t}$\\
Prediction Model: $F_1,F_2$\\

\STATE Initialize model parameters ${\mathbf \theta}_1,{\mathbf \theta}_2$
%\FOR{$N_1$}
\FOR{$i = 1$ to $N_1$}
\STATE{Get random minibatch $\mathcal{D}_s^{\prime}$
from $\mathcal{D}_s$.}
\STATE Update model parameters ${\mathbf \theta}_j (j=1,2)$ by ascending their stochastic gradients with respect to Eq.~\ref{eq:loss}:
\begin{center}
{$\nabla_{{\mathbf \theta}_j}L_{s_j}$}
\end{center}
\ENDFOR
%\FOR 
\FOR {$i=N_1+1$ to $N_1+N_3$}
\IF {$i \% N_2 == 0$}
\STATE Update the label distribution $\mathbf{w}$ via Eq.~\ref{eq:weight}.
%\begin{center}
%$\mathbf{w}=\frac{1}{2n_t}\sum_{i=1}^{n_t}\left(p_1(\mathbf{y}|\mathbf{x})+p_2(\mathbf{y}|\mathbf{x})\right)$
%\end{center}
\ENDIF
\STATE{Get random minibatch $\mathcal{D}^{\prime}$
from $\mathcal{D}_s$ and $\mathcal{D}_t$.}
\STATE Update model parameters $\theta_j (j=1,2)$ by ascending their stochastic gradient with respect to Eq.~\ref{eq:loss_all}:
\begin{center}
$\nabla_{{\mathbf \theta}_j}L_\mathrm{total}$
%$\nabla_{{\mathbf \theta}_j}\frac{1}{|\mathcal{D}^{\prime}|}\sum_{{\mathbf x}_i^s,{\mathbf x}_i^t,y_i^s\in\mathcal{D}^{\prime}}L_\mathrm{total}(\mathbf{ x}_i^s,\mathbf{x}_i^t,y_i^s)$
\end{center}
\ENDFOR

\end{algorithmic}
\end{algorithm}

\subsection{Training Procedure}\label{sec:procedure}
We summarize how we integrated the two ideas to achieve partial domain adaptation. The complete training procedure is summarized in Alg.~\ref{alg:twins}. Phase~2 and Phase~3 are also visualized in Fig.~\ref{fig:training}.

\noindent{\textbf{Phase~1: Pre-train classifiers with only source samples.}}\ \ We pre-train the two networks by using source samples. In the training phase, we do not use any adaptation methods. The phase is not required when we possess access to the pre-trained model.\par
\noindent{\textbf{Phase~2: Estimate the target label distribution.}}\ \ We estimate the label distribution of the target by using two classification networks (Eq.~\ref{eq:weight}) using all training target samples.\par
\noindent{\textbf{Phase~3: Optimize the parameters of classifiers.}}\ \ We train networks by using weighted classification loss on source samples (Eq.~\ref{eq:loss_weighted}) and inconsistency loss calculated on target samples (Eq.~\ref{eq:loss_t}). The overall loss function used in the phase is as follows:
\begin{equation}
    L_{\mathrm{total}}=L_s^{\prime}+L_t.
     \label{eq:loss_all}
\end{equation}\par

\noindent{\textbf{Repeat Phase~2 and Phase~3 alternately.}}\ \ We train the two networks by repeating Phase~2 and Phase~3 alternately. In Phase~3, the target features are aligned with the source, and this ensures that the estimation of the label distribution is more accurate. This also results in better feature alignment. Therefore, the label distribution estimation and feature alignment by using the estimated label distribution should benefit from each other. Thus, we alternately change Phase~2 and Phase~3. The training objective is simplified when compared to MCD~\cite{MCD} since we do not employ adversarial learning. Although the phase of estimating the target label distribution is required, the objective to train the network involves simply minimizing Eq.~\ref{eq:loss_all}.

\subsection{Theoretical Insight}\label{sec:theory}
Given that MCD~\cite{MCD} is motivated by the theory proposed by Ben-David \etal~\cite{ben2010theory} and that the proposed method is related to it, our aim involves demonstrating the relationship between our method and the theory in this section. Ben-David \etal~\cite{ben2010theory} proposed a theory that bounds the expected error on the target samples $R_{\mathcal{T}}(h)$ by using the following three terms: (i) the expected error on the source domain $R_{\mathcal{S}}(h)$; (ii) the $\mathcal{H} \Delta \mathcal{H}$-distance ($d_{{\hdistance}}(\mathcal{S},\mathcal{T})$), which is measured as the discrepancy between two classifiers; and (iii) the shared error of the ideal joint hypothesis $\lambda$ that is considered as a constant value. Specifically, $\mathcal{S}$ and $\mathcal{T}$ denote source and target domains, respectively.
\begin{theorem1}\label{th:th_1}
  Let $H$ be the hypothesis class. Given two domains $\mathcal{S}$ and $\mathcal{T}$, we obtain the following:
   \begin{eqnarray}
  \forall h \in H, R_{\mathcal{T}}(h)\leq R_{\mathcal{S}}(h)  +\frac{1}{2}{d_{\mathcal{H} \Delta \mathcal{H}}(\mathcal{S},\mathcal{T})}+\lambda,
  \label{eq:main}
  \end{eqnarray}
where
\begin{align}
      &\frac{1}{2}d_{{\hdistance}}(\mathcal{S},\mathcal{T})\\\nonumber
      =&\sup_{(h,h{'})\in \mathcal{H}^{2}}\left| \underset{\mathbf{x}\sim \mathcal{S}}{\mathbf{E}}{\rm I}\bigl[h(\mathbf{x})\!\neq\! h^{'}(\mathbf{x}) \bigr]\mathalpha{-} \underset{\mathbf{x}\sim \mathcal{T}}{\mathbf{E}} {\rm I}\bigl[h(\mathbf{x})\!\neq\! h^{'}(\mathbf{x}) \bigr]\right|, \\
&\lambda=\min \left[R_{\mathcal{S}}(h)+R_{\mathcal{T}}(h)\right].
\end{align}
Here, $R_{\mathcal{T}}(h)$ denotes the error of hypothesis $h$ on the target domain, and $R_{\mathcal{S}}(h)$ denotes the corresponding error on the source domain. Additionally, ${\rm I}[a]$ denotes the indicator function, and this corresponds to 1 if the predicate a is true and 0 otherwise.
\label{th:thm1}
\end{theorem1}\par
Based on the theory, we can argue that it is possible to approximate the divergence between two domains by using the discrepancy between two classifiers.
In a study of MCD \cite{MCD}, the aim involved approximating $d_{{\hdistance}}(\mathcal{S},\mathcal{T})$ by adversarial training between two classifiers and a feature extractor. They assume that the term $\scalebox{0.9}{$\displaystyle \underset{\mathbf{x}\sim \mathcal{S}}{\mathbf{E}} {\rm I}\bigl[h(\mathbf{x}) \neq h^{'}(\mathbf{x}) \bigr]$}$ is extremely low because the source samples are labeled. Therefore, $d_{{\hdistance}}(\mathcal{S},\mathcal{T})$ is approximately calculated as follows: $\scalebox{0.9}{$\displaystyle\sup_{(h,h{'})\in \mathcal{H}^{2}}\underset{\mathbf{x}\sim \mathcal{T}}{\mathbf{E}} {\rm I}\bigl[h(\mathbf{x}) \neq h^{'}(\mathbf{x}) \bigr],$}$ and this denotes the supremum of the expected discrepancy of two classifiers' predictions on target samples. To calculate the maximum of the discrepancy, they train the two classification networks to disagree on their predictions on the target.\par
%In our study, 
We minimize the left side of the following inequality:
%\begin{equation}
\scalebox{0.9}{
$\displaystyle
\underset{{\mathbf{x} \sim \mathcal{T}}}{\mathbf{E}}{\rm I}\bigl[h(\mathbf{x})\neq h^{'}(\mathbf{x}) \bigr]
\leq\sup_{(h,h{'})\in \mathcal{H}^{2}}\underset{\mathbf{x}\sim \mathcal{T}}{\mathbf{E}}{\rm I}\bigl[h(\mathbf{x})\neq h^{'}(\mathbf{x})\bigr].
$
}
%\end{equation}
Therefore, we approximate the divergence between domains lower than the divergence used in MCD~\cite{MCD}. If we can completely estimate the label distribution of the target domain in Phase~2, then strictly aligning feature distribution by using $\scalebox{0.9}{$\displaystyle \sup_{(h,h{'})\in \mathcal{H}^{2}}\underset{\mathbf{x}\sim \mathcal{T}}{\mathbf{E}} {\rm I}\bigl[h({\bf x}) \neq h^{'}(\mathbf{x}) \bigr]$}$ is effective. However, in reality, the estimation always includes a few errors, and thus minimizing the relaxed divergence $\scalebox{0.9}{$\displaystyle \underset{{\mathbf{x} \sim \mathcal{T}}}{\mathbf{E}} {\rm I}\bigl[h(\mathbf{x}) \neq h^{'}(\mathbf{x}) \bigr]$}$ is considered as appropriate.

% \begin{figure}[t]
% \centering
%   \includegraphics[clip,width=0.9\linewidth]{figure/data.eps}
% \caption{Examples of images in each dataset. Left: Digits and traffic sign datasets. Right: \textit{Office-31}.}
% \label{fig:data}
% \end{figure}

\section{Experiments}
We evaluate our method on several datasets to compare our method with state-of-the-art deep learning methods for domain adaptation. It should be noted that all experiments are performed in the unsupervised setting where labels in the target domain are not given. The goal of the experiments involves demonstrating that our method is effective on both digits classification and general object classification datasets.%We explain each setting, show the results, and discuss the analysis of our method by using the datasets.

{\tabcolsep = 0.6mm
\begin{table}[t]
\centering
\scalebox{0.9}{
\begin{tabular}{l|cccc}\hline
\toprule[1.5pt]
Method 
& 
\begin{tabular}{c}
MNIST\\$\downarrow$\\USPS
\end{tabular}
& 
\begin{tabular}{c}
USPS\\$\downarrow$\\MNIST
\end{tabular}
& 
\begin{tabular}{c}
SVHN\\$\downarrow$\\MNIST
\end{tabular}
& 
\begin{tabular}{c}
SYN SIGNS\\$\downarrow$\\GTSRB
\end{tabular} 
\\ \hline

Source Only & 85.2 & 80.0 & 73.9 & 89.2 \\ \hline
\multicolumn{5}{c}{\textit{Methods for Unsupervised Domain Adaptation}} \\\hline
DAN~\cite{DAN} & 83.5 & 80.7& 70.9 & 90.2 \\
DANN~\cite{GRL} & 67.1 & 72.1 & 39.8 & 55.5 \\
MCD~\cite{MCD} & 66.4 & 59.4 & 71.2 & 93.1 \\
\hline
 \multicolumn{5}{c}{\textit{Methods for Partial Domain Adaptation}} \\\hline
IWAN~\cite{ImportanceWA} & 90.6 & 85.7 &75.6 & 77.7\\
PADA~\cite{PADA} & 78.2 & 73.9 & 44.1 & 71.2\\ \hline
TWINs (Ours)& \textbf{96.3} & \textbf{90.2} & \textbf{99.6} & \textbf{95.5}\\ 
\bottomrule[1.5pt]
\end{tabular}
}
\caption{Accuracy for digits and traffic sign datasets. In the tasks of MNIST $\to$ USPS, USPS $\to$ MNIST, and SVHN $\to$ MNIST, the source domain includes 10 classes, and the target domain includes 5 classes. In the task of SYN SIGNS $\to$ GTSRB, the source domain includes 43 classes, and the target domain includes 20 classes. TWINs achieves strongest results on all four evaluated partial domain adaptation scenarios.}
\label{tab:digits}
\end{table}
}

% \begin{table}[t]
% \centering
% \begin{tabular}{l|cccc|c}\hline
% \toprule[1.5pt]
% Method 
% & 
% MNIST $\rightarrow$ USPS
% & 
% USPS $\rightarrow$ MNIST
% & 
% SVHN $\rightarrow$ MNIST
% & 
% SYN SIGNS $\rightarrow$ GTSRB
% & Avg.
% \\ \hline

% Source Only & 85.2 & 80.0 & 73.9 & 89.2 & 82.1\\ \hline
% \multicolumn{6}{c}{\textit{Methods for Unsupervised Domain Adaptation}} \\\hline
% DAN~\cite{DAN} & 83.5 & 80.7& 70.9 & 90.2 & 81.3 \\
% DANN~\cite{GRL} & 67.1 & 72.1 & 39.8 & 55.5 & 58.6\\
% MCD~\cite{MCD} & 66.4 & 59.4 & 71.2 & 93.1 & 72.5\\
% \hline
%  \multicolumn{6}{c}{\textit{Methods for Partial Domain Adaptation}} \\\hline
% IWAN~\cite{ImportanceWA} & 90.6 & 85.7 &75.6 & 77.7 & 82.4\\
% PADA~\cite{PADA} & 78.2 & 73.9 & 44.1 & 71.2 & 66.8\\ \hline
% TWINs (Ours)& \textbf{96.3} & \textbf{90.2} & \textbf{99.6} & \textbf{95.5} & \textbf{95.4}\\ 
% \bottomrule[1.5pt]
% \end{tabular}

% \caption{Accuracy for digits and traffic sign datasets. In the tasks of MNIST $\to$ USPS, USPS $\to$ MNIST, and SVHN $\to$ MNIST, the source domain includes 10 classes, and the target domain includes 5 classes. Conversely, in the task of SYN SIGNS $\to$ GTSRB, the source domain includes 43 classes, and the target domain includes 20 classes. Specifically, TWINs achieves strongest results on all four evaluated partial domain adaptation scenarios.}
% \label{tab:digits}
% \end{table}

\subsection{Experiments on Digit and Traffic Sign Datasets}\label{sec:digits}
In the experiment, we evaluate our proposed method with respect to adaptation for digits and traffic sign datasets. The networks are trained from scratch in the setting.\par
\noindent{\textbf{Setup.}}\ \ We utilize digit datasets (including MNIST~\cite{mnist}, Street View House Numbers (SVHN)~\cite{SVHN}, and US Postal handwritten digit dataset (USPS)~\cite{USPS}) and traffic sign datasets (including Synthetic Traffic Signs (SYN SIGNS)~\cite{SYN_SIGNS} and German Traffic Signs Recognition Benchmark (GTSRB)~\cite{GTSRB}).
Digit datasets consist of 10 classes, and traffic sign datasets consist of 43 classes, respectively.
%We provide examples of images contained in each dataset in the left of Figure~\ref{fig:data}. 
We evaluate our method across four domain adaptation tasks (\ie, \textbf{MNIST $\to$ USPS}, \textbf{USPS $\to$ MNIST}, \textbf{SVHN $\to$ MNIST}, and \textbf{SYN SIGNS $\to$ GTSRB}). When the dataset is used as the target domain, we use the first five classes in the experiment on digit datasets and the first twenty classes in the experiment on traffic sign datasets in ascending order, and we use all images in the classes for training.\par
Extant studies do not report the results for the aforementioned adaptation scenarios of PDA, and thus we employ the optimization procedure and CNN architecture used in~\cite{MCD}, basically. Optimization proceeds via the Adam optimizer~\cite{Adam} for $30$ epochs with a learning rate of $2.0 \times 10^{-4}$, a $\beta_1$ of 0.9, a $\beta_2$ of 0.999, and a batch size of $256$ images (128 per domain) in all experiments. We pre-train our classifiers by only source samples in the first 10 epochs (Phase~1), and we subsequently estimate the label distribution of target samples (Phase~2) and optimize the parameters of classifiers by source and target samples (Phase~3), repeatedly.
%The label distribution estimation is performed at each epoch after the first 10 epochs. 
We follow the protocol of unsupervised domain adaptation by using all labeled source data and all unlabeled target data and do not use validation samples to tune hyperparameters per each adaptation scenario. Further details are provided in our supplementary material due to space limitations.\par
\noindent{\textbf{Results.}}\ \ Our method achieves better accuracy on all four partial domain adaptation scenarios as shown in Tab.~\ref{tab:digits}. 
Our method outperforms existing methods for PDA. The results indicate the effectiveness of using task-specific classifier's inconsistency as the distance between domains. IWAN~\cite{ImportanceWA} and PADA~\cite{PADA} do not exhibit better performance even compared with the methods for UDA. This is potentially because the models are based on DANN wherein the training process is unstable in a few scenarios~\cite{ADDA}. Furthermore, they propose to partially align the feature distributions by controlling weight on source samples in training a domain classifier. This can make the training of the domain classifier more unstable.\par

\begin{figure}[t] 
\centering
\subfigure[Source only]{\includegraphics[trim=0cm 3.5cm 0cm 3.5cm,clip,width=0.45\linewidth]{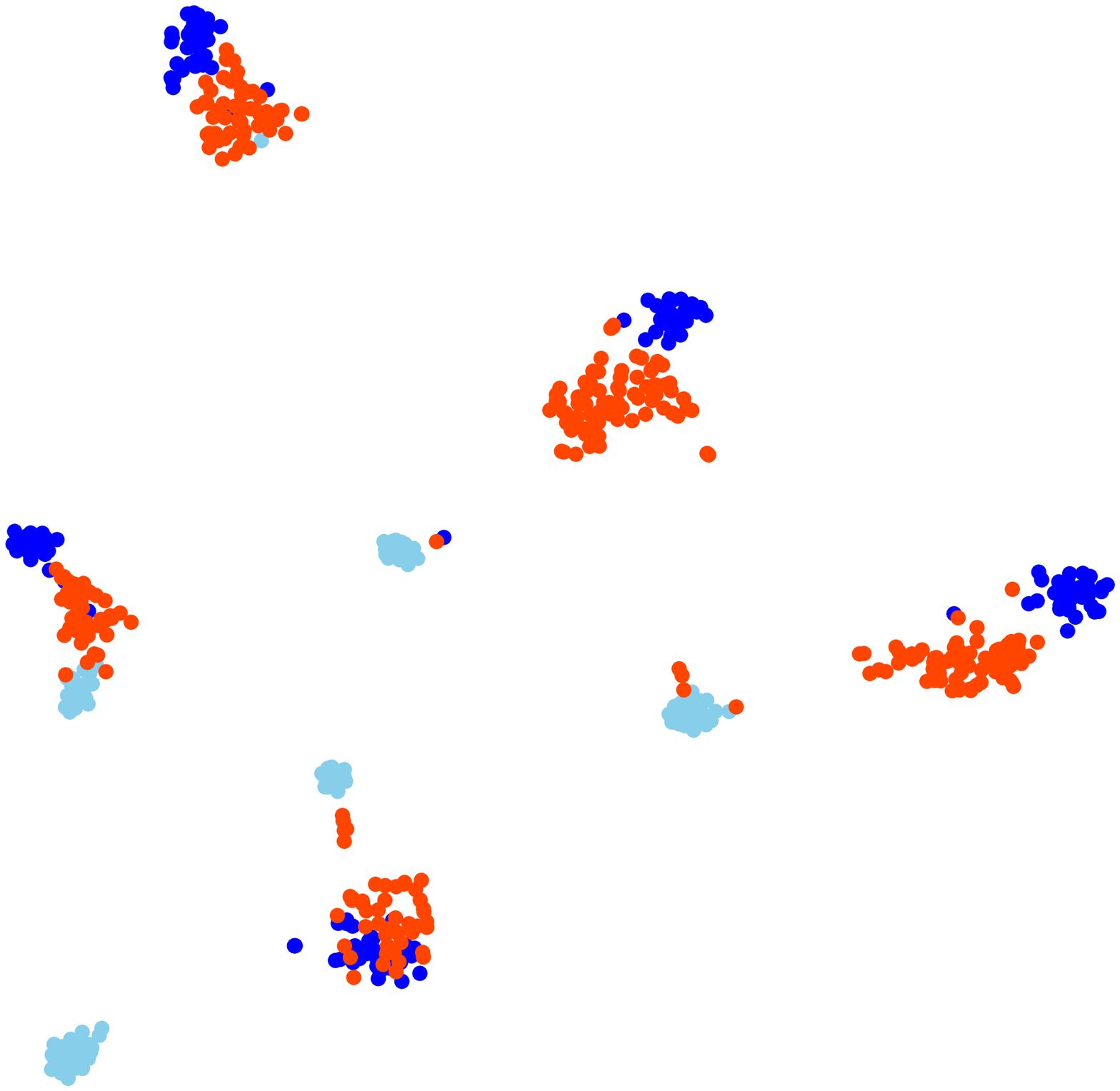}
\label{fig:source_emb}}
\subfigure[MCD]{\includegraphics[trim=0cm 3.5cm 0cm 3.5cm,clip,width=0.45\linewidth]{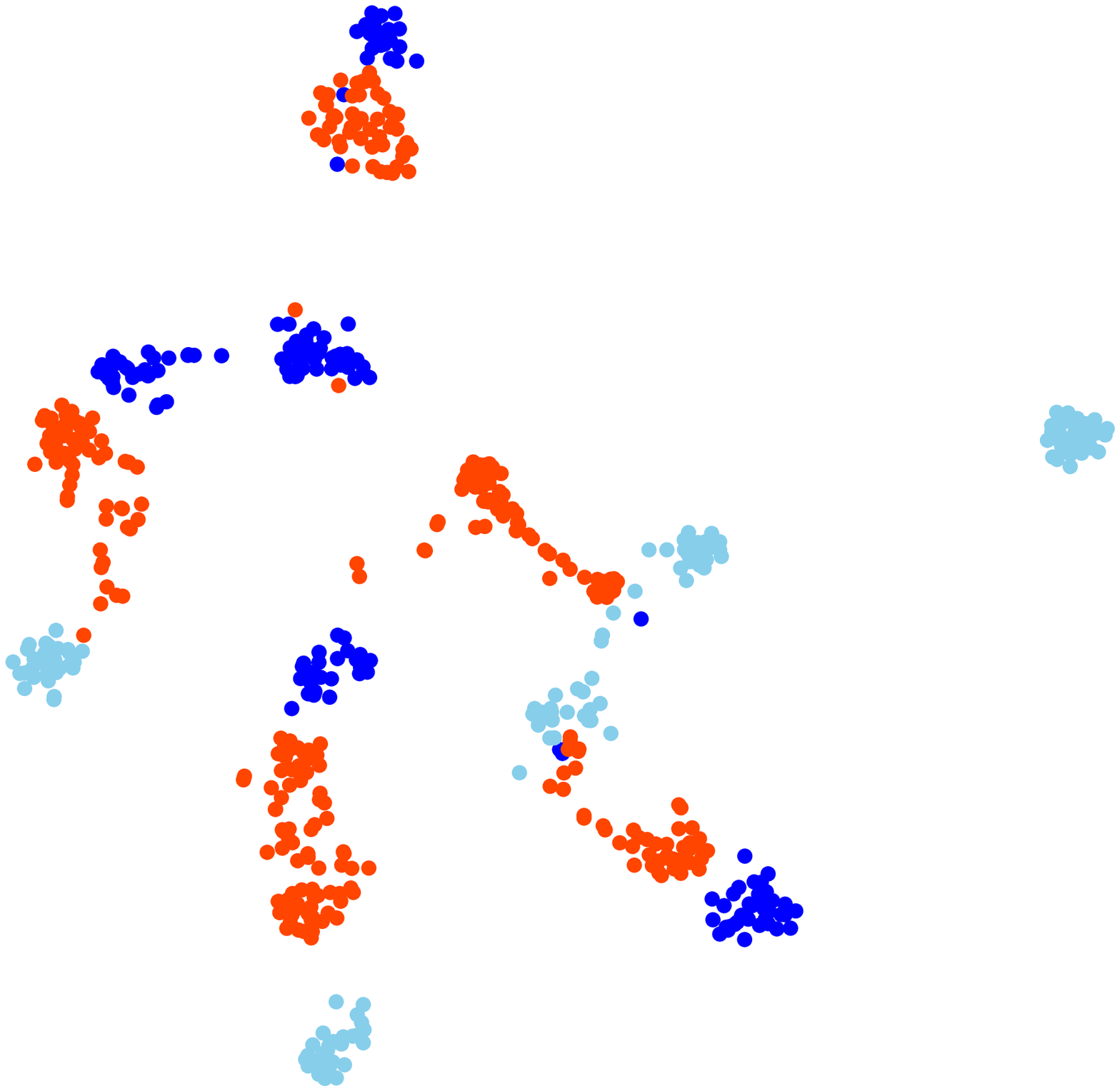}
\label{fig:MCD_emb}}
\subfigure[PADA]{\includegraphics[trim=0cm 3.5cm 0cm 3.5cm,clip,width=0.45\linewidth]{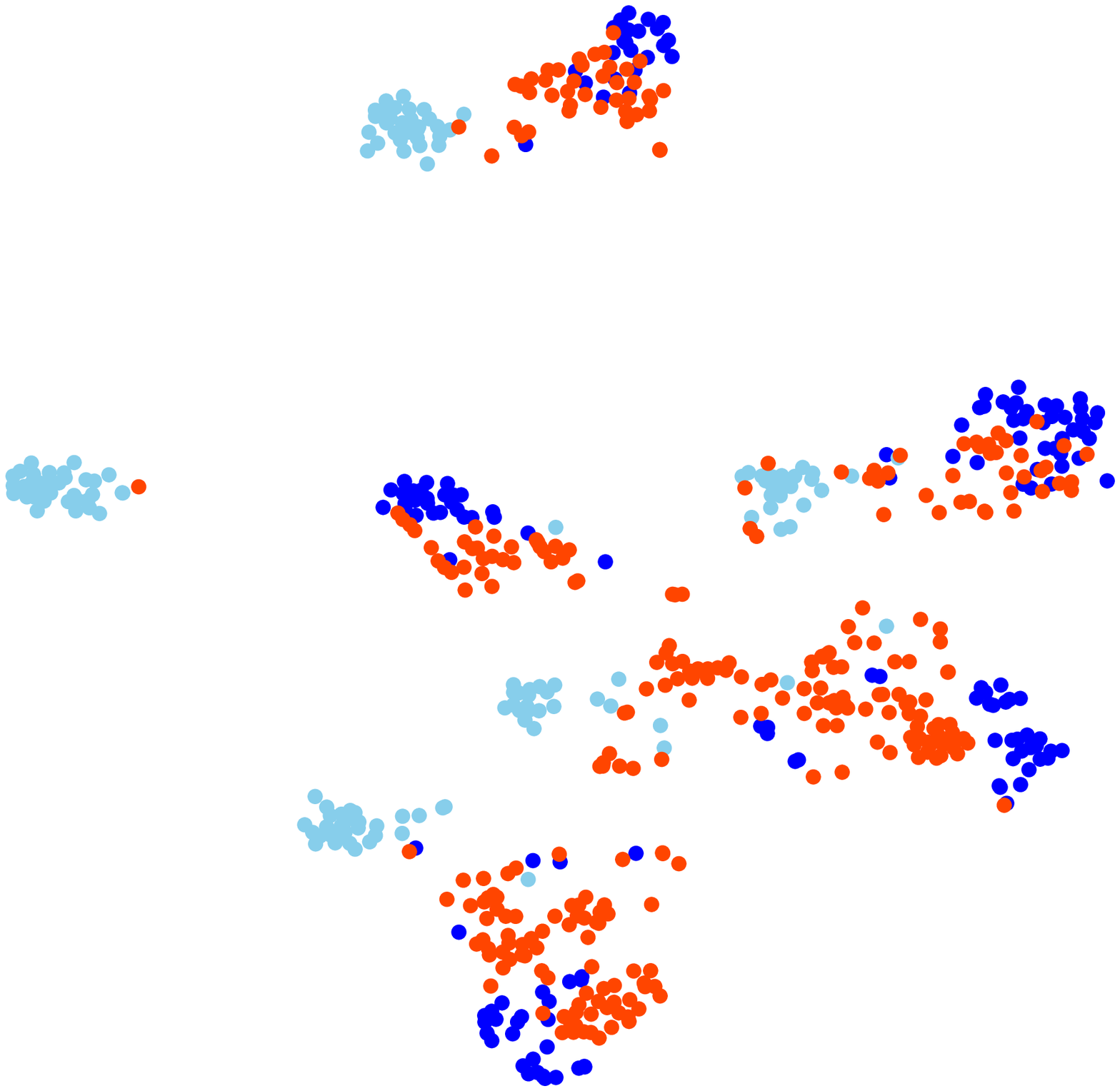}\label{fig:PADA_emb}}
\subfigure[TWINs]{\includegraphics[trim=0cm 3.5cm 0cm 3.5cm,clip,width=0.45\linewidth]{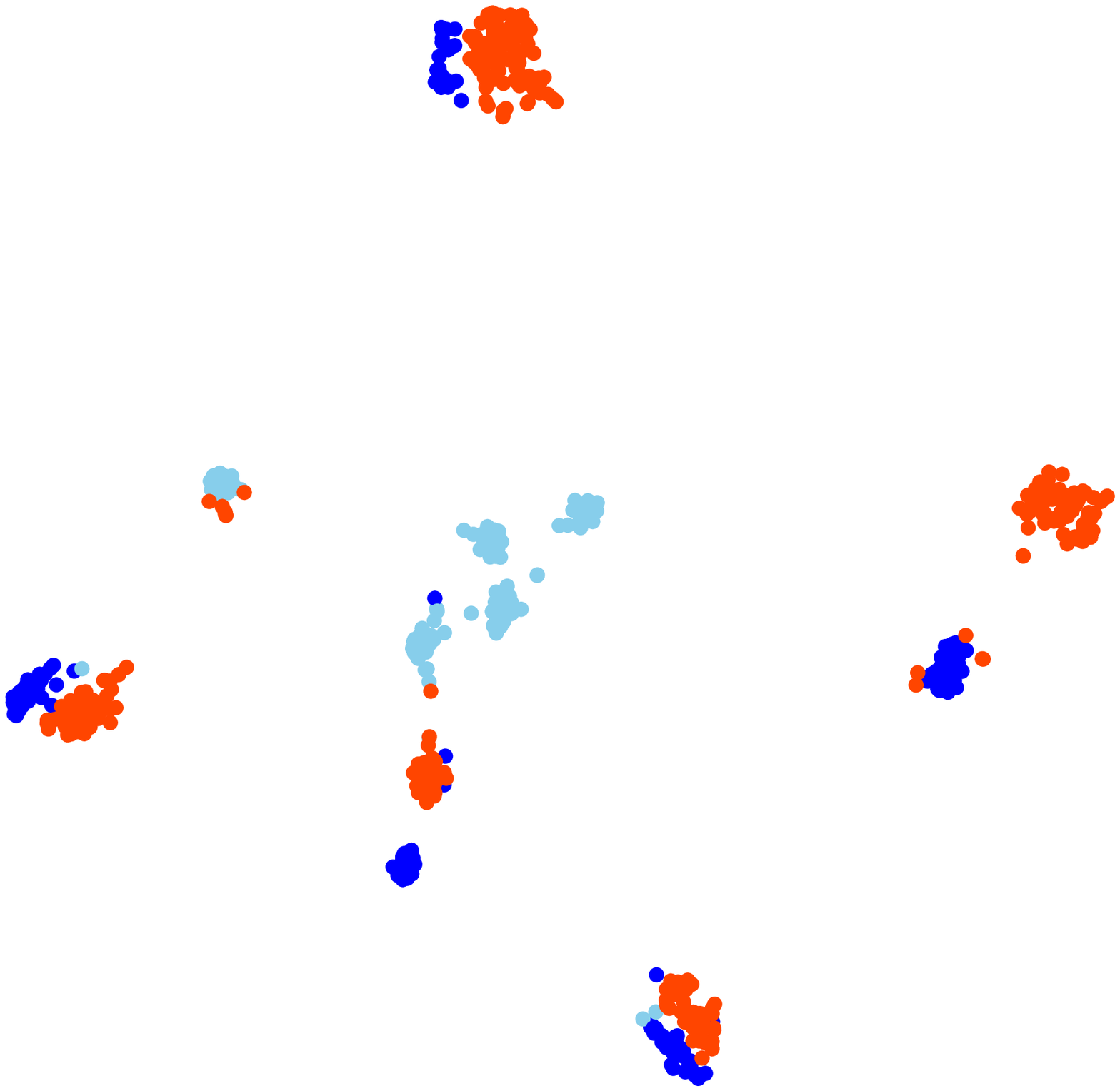}
\label{fig:twins_emb}}
\caption{T-SNE visualization of features obtained from the second last fully connected layer of (a) source only, (b) MCD, (c) PADA, and (d) TWINs. The transfer task is MNIST (10 classes) $\to$ USPS (5 classes). Blue/light blue dots correspond to the source domain samples in which the classes are present/absent in the target domain while orange dots correspond to the target domain samples. All samples are testing samples. The results indicate our method enable target samples to be aligned with the classes present in the target domain. Furthermore, our method extracts discriminative features considering classification boundaries.}
\label{fig:t_SNE}
\end{figure}

\begin{table*}[t!]
\centering
\scalebox{0.9}{
\begin{tabular}{l|cccccc|c}
\toprule[1.5pt]
Method 
& \textbf{A $\to$ W} & \textbf{D $\to$ W} & \textbf{W $\to$ D} & \textbf{A $\to$ D} & \textbf{D $\to$ A} & \textbf{W $\to$ A} & Avg.\\ \hline

ResNet~\cite{ResNet} & 54.5 & 94.6 & 94.2 & 65.6 & 73.2 & 71.7 & 75.6\\ \hline
 \multicolumn{8}{c}{\textit{Methods for Unsupervised Domain Adaptation}} \\\hline

DAN~\cite{DAN} & 46.4 & 53.6 & 58.6 & 42.7 & 65.7 & 65.3 & 55.4\\
DANN~\cite{GRL} & 41.4 & 46.8 & 38.9 & 41.4 & 41.3 & 44.7 & 42.4\\
ADDA~\cite{ADDA} & 43.7 & 46.5 & 40.1 & 43.7 & 42.8 & 46.0 & 43.8\\
RTN~\cite{RTN} & 75.3 & 97.1 & 98.3 & 66.9 & 85.6 & 85.7 & 84.8\\
JAN~\cite{JAN} & 43.5 & 53.6 & 41.4 & 35.7 & 51.0 & 51.6 & 46.1\\
LEL~\cite{LEL} & 73.2 & 93.9 & 96.8 & 76.4 & 83.6 & 84.7 & 84.8\\
\hline
 \multicolumn{8}{c}{\textit{Methods for Partial Domain Adaptation}} \\\hline
IWAN~\cite{ImportanceWA} & 77.7 & 98.4 & \bf{100} & 81.5 & 77.7 & 73.4 & 84.8\\
PADA~\cite{PADA} & \bf{86.5} & \bf{99.3} & \bf{100} & 82.2 & 92.7 & \bf{95.4}& 92.7\\ \hline
TWINs (Ours) & 86.0 & \textbf{99.3} & \textbf{100} & \textbf{86.8} & \textbf{94.7} & 94.5 & \textbf{93.6} \\ 
\bottomrule[1.5pt]
\end{tabular}
}
\caption{Accuracy on \textit{Office-31} dataset. The source domain includes 31 classes, and the target domain includes 10 classes. TWINs achieves results that either equal or surpass those of existing methods.}
\label{tab:office}
\vspace{-1ex}
\end{table*}

\noindent\textbf{Feature Visualization.}\ \ We visualize feature distribution via t-SNE~\cite{t_SNE} to qualitatively compare our method with other methods such as MCD~\cite{MCD} and PADA~\cite{PADA}. The features are extracted from the middle layer of the network.
The visualized feature distribution is shown in Fig.~\ref{fig:t_SNE}.
As shown in Fig.~\ref{fig:twins_emb}, our method aligns target samples with source classes present in the target domain and acquires discriminative features by considering the task-specific decision boundary, thereby enabling a high performance classification.
When MCD is applied (Fig.~\ref{fig:MCD_emb}), target samples are not correctly aligned with source classes present in the target domain. Furthermore, they fail to extract discriminative features for target samples because their aim involves matching the overall feature distribution.
As shown in Fig.~\ref{fig:PADA_emb}, PADA extracts ambiguous features for target samples and fails to align target samples to the source classes present in the target domain. Comparing this result with ours, we can see the effectiveness of considering decision boundary's information for the feature alignment.
\par

\subsection{Experiments on \textbf{\textit{Office-31}} Datasets}
We further evaluate our proposed method on object classification task.\par
\noindent{\textbf{Setup.}}\ \ \textit{Office-31}~\cite{office} is a benchmark dataset for domain adaptation. 
It contains a total of 4110 images across 31 classes in the following three domains: \textit{Amazon} (\textbf{A}) that contains of images from the web downloaded from online merchants (\url{www.amazon.com}), and \textit{Webcam} (\textbf{W}) and \textit{DSLR} (\textbf{D}) that are captured with a web camera and a digital SLR camera, respectively.
%We present the examples of images contained in each domain in the right of Fig.~\ref{fig:data}. 
We evaluate our method across the following six scenarios: \textbf{A $\to$ W}, \textbf{D $\to$ W}, \textbf{W $\to$ D}, \textbf{A $\to$ D}, \textbf{D $\to$ A}, and \textbf{W $\to$ A}. When a domain is used as the target domain, we use the samples of ten classes shared by \textit{Office-31} and Caltech-256~\cite{caltech} by following~\cite{PADA,ImportanceWA}.\par
We follow standard evaluation protocols and use all labeled source data and all unlabeled target data for unsupervised domain adaptation. We use PyTorch-provided models of ResNet-50~\cite{ResNet} pre-trained on ImageNet~\cite{ImageNet} as two classifiers of our method. Essentially, we use the same hyperparameters and network architectures as used in~\cite{PADA}. The finally fully connected layer is removed and replaced by a three-layered fully connected network that is randomly initialized. We fine-tune all pre-trained feature layers and train the initialized fully connected layer. We use mini-batch stochastic gradient descent (SGD) with a momentum of 0.9, and the learning rate is adjusted during SGD by using $\eta_p = \frac{\eta_0}{(1+\alpha p)^\gamma}$ where $p$ denotes the training progress changing from 0 to 1, while $\eta_0=0.001$, $\alpha=0.001$, and $\gamma=0.75$. 
In the experiment, we possess access to the pre-trained model, and thus we do not use Phase~1 and begin training a model from Phase~3.\par

\noindent{\bf Results.}\ \ Our method achieves results that either equal or surpass those of existing methods as shown in Tab.~\ref{tab:office}. In addition to experiments on digits and traffic sign datasets, methods for UDA are also prone to exhibit a worse performance than that of model trained only source samples. Methods for PDA perform well when using a pre-trained CNN feature extractor and fine-tuning it. However, the performance of our model exceeds that of extant methods on average because our method accounts for the relationship between target samples and the task-specific decision boundary.

\subsection{Empirical Analysis}\label{sec:empirical}
We conduct empirical analyses to clarify the characteristic of our method.
%In this section, we present the study on the number of target classes, the estimated label distribution of target samples, and the ablation study.
\par
\begin{figure*}[t]
\begin{center}
\includegraphics[width=\linewidth]{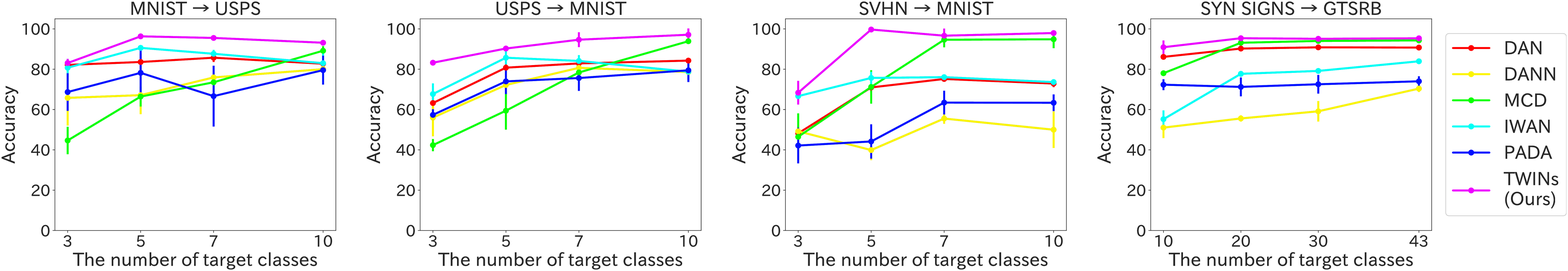}
\caption{Accuracy when the number of target classes varies. From left to right, the adaptation scenarios are MNIST $\to$ USPS, USPS $\to$ MNIST, SVHN $\to$ MNIST, and SYN SIGNS $\to$ GTSRB. 
Our method performs better than other methods in all settings.}
\label{fig:class_number}
\end{center}
\vspace{-2ex}
\end{figure*}

% \begin{figure}[t]
% \begin{center}
% \subfigure[MNIST$\to$USPS]{\includegraphics[width=0.48\linewidth]{figure/mnist2usps.eps}\label{fig:mnist2usps}
% }
% \subfigure[USPS$\to$MNIST]{\includegraphics[width=0.48\linewidth]{figure/usps2mnist.eps}\label{fig:usps2mnist}
% }
% \subfigure[SVHN$\to$MNIST]{\includegraphics[width=0.48\linewidth]{figure/svhn2mnist.eps}\label{fig:svhn2mnist}
% }
% \subfigure[SYN SIGNS$\to$GTSRB]{\includegraphics[width=0.48\linewidth]{figure/syn2gtsrb.eps}\label{fig:syn2gtsrb}
% }
% \caption{Accuracy when the number of target class changes. %in the task corresponding to (a) MNIST $\to$ USPS, (b) USPS $\to$ MNIST, (c) SVHN $\to$ MNIST, (d) SYN SIGNS $\to$ GTSRB. 
% Our method also outperforms other methods when the number of target class is low.}
% \label{fig:class_number}
% \end{center}
% \end{figure}

\noindent{\bf Study on the Number of Target Classes.}\ \ We explore the prediction performance on our method when the number of target classes varies. We list the experimental results %when the number of target classes varies 
in the task of MNIST $\to$ USPS, USPS $\to$ MNIST, SVHN $\to$ MNIST, and SYN SIGNS $\to$ GTSRB. Fig.~\ref{fig:class_number} shows the results. In all the settings, the performance of our method is comparable to or exceeds that of other methods. When the number of target classes is equivalent to source classes, which is the standard unsupervised domain adaptation setting, our method also performs better than other methods. Therefore, our method is useful for the standard domain adaptation setting too.
When the number of target classes is small, the performance of our method occasionally drops although it still performs better than other existing methods.
\par

\begin{figure}[t]
\begin{center}
\subfigure[Ground Truth]{\includegraphics[width=0.48\linewidth]{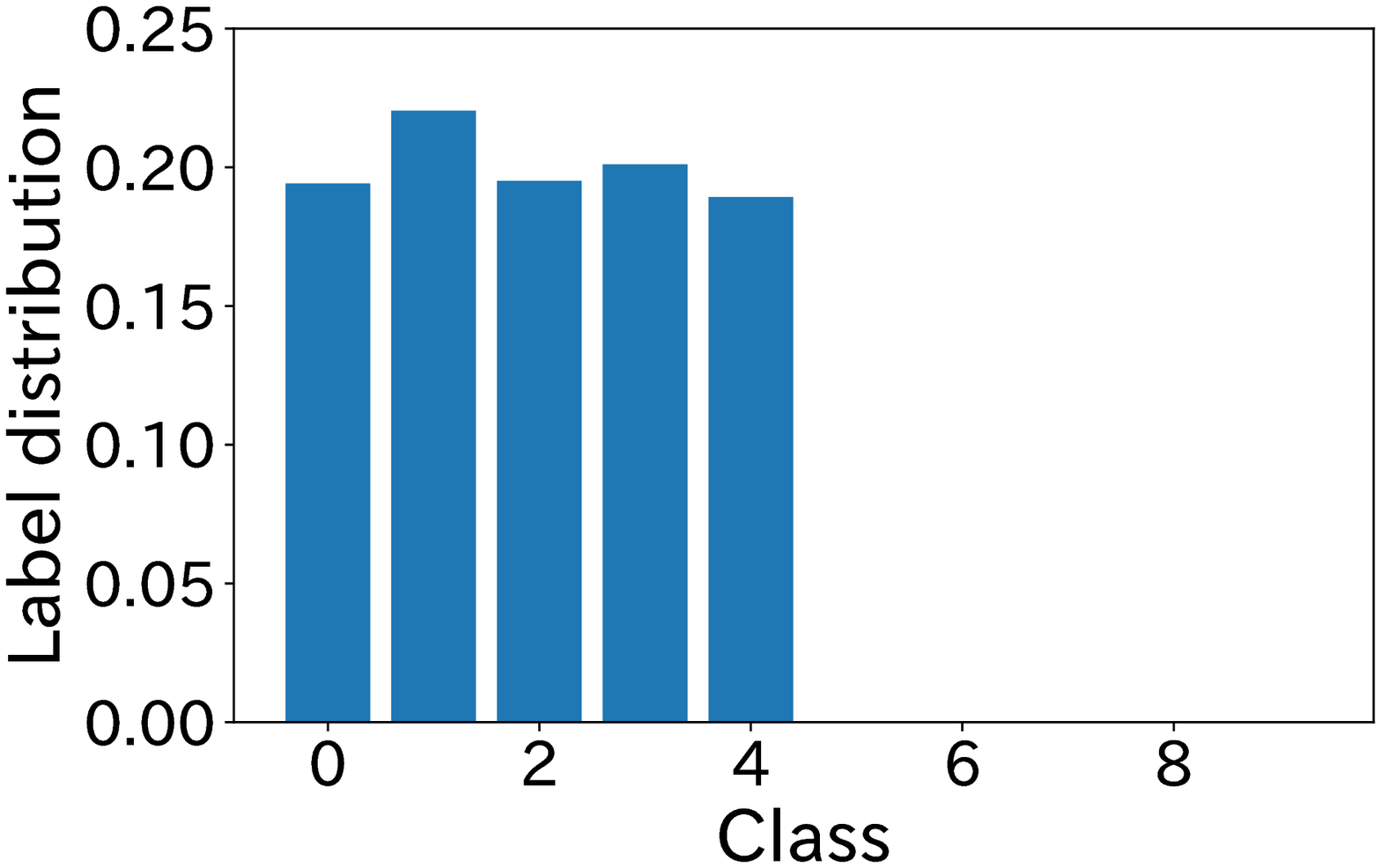}\label{fig:ground_truth}
}
\subfigure[Phase 1]{\includegraphics[width=0.48\linewidth]{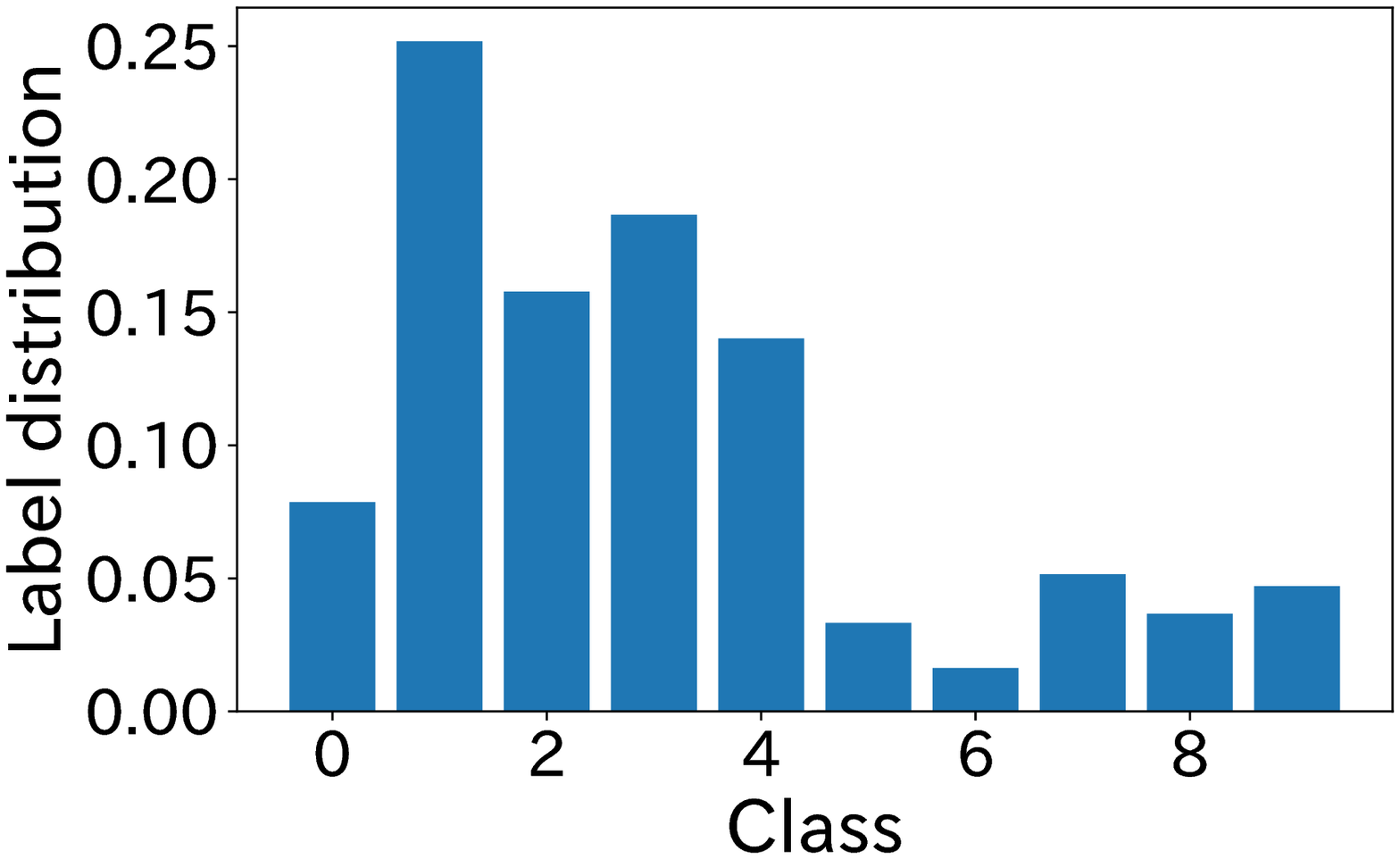}\label{fig:phase1}
}
\subfigure[11 epoch]{\includegraphics[width=0.48\linewidth]{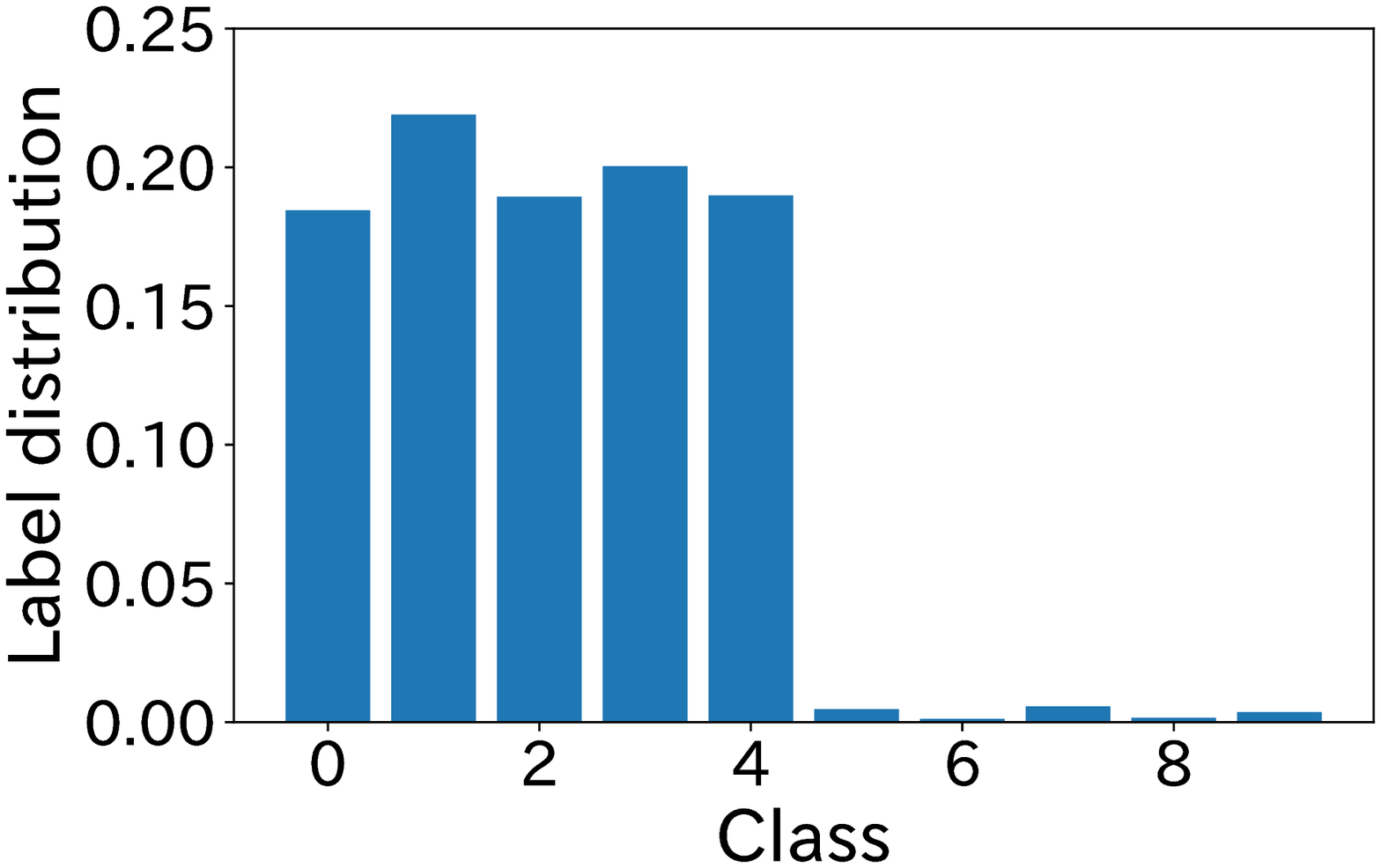}\label{fig:middle}
}
\subfigure[30 epoch]{\includegraphics[width=0.48\linewidth]{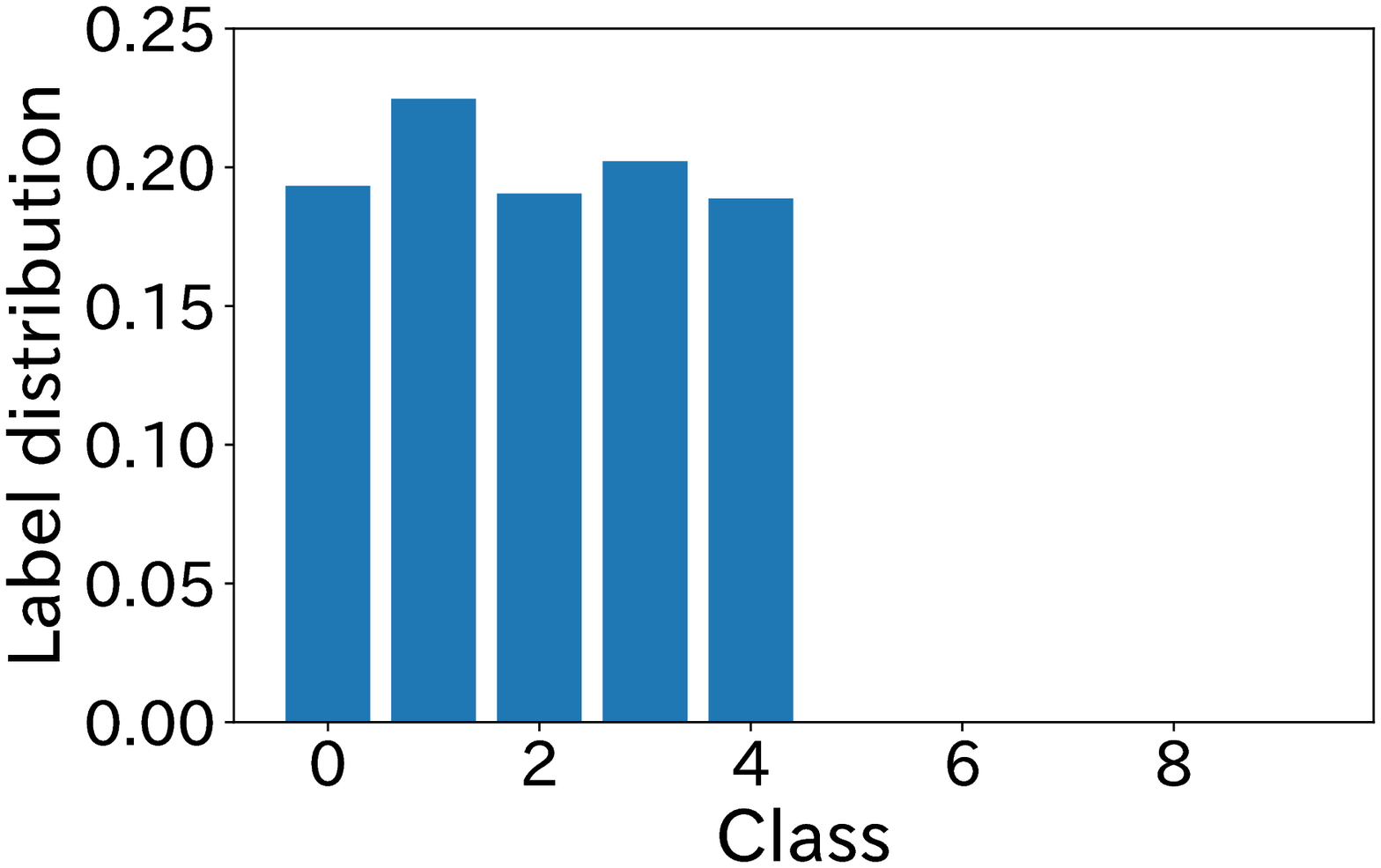}\label{fig:final}
}
\caption{The estimated label distribution of target samples. (a) Ground Truth shows the real label distribution of the target samples. The estimated class distributions (b) after Phase~1, (c) after 11 epoch, and (d) after 30 epoch are shown here.}
\label{fig:class_distribution}
\end{center}
\vspace{-2ex}
\end{figure}

\noindent{\bf Estimated Label Distribution of Target Samples.}\ \ Our method estimates the label distribution of target samples to align them with only the source classes present in the target domain. The estimated class distribution on the task SVHN (10 classes) $\to$ MNIST (5 classes) is shown Fig.~\ref{fig:class_distribution}. Fig.~\ref{fig:ground_truth} shows the true class distribution of training samples in the target domain. Fig.~\ref{fig:phase1} shows the estimated label distribution of the target domain after finishing pre-training models with only source samples (Phase~1). This is prior to the adaptation, and thus the estimated label distribution is far from the real distribution and is assigned to the class absent in the target domain. Fig.~\ref{fig:middle} shows the estimated distribution one epoch after starting Phase~3 (11~epoch). The estimated class distribution gets closer to the ground truth although a few samples are assigned to the class absent in the target domain. Fig.~\ref{fig:final} shows the estimated distribution after completing the series of training procedure (30~epoch). All target samples are aligned with source classes present in the target domain, and the distribution is closer to the ground truth.
\par

\begin{table}[t]
\begin{center}
\scalebox{0.88}{
\begin{tabular}{l|llll}
\toprule[1.5pt]
Method & $N\mathalpha{=}3$ & $N\mathalpha{=}5$ & $N\mathalpha{=}7$ & $N\mathalpha{=}10$ \\ \hline
 &\multicolumn{4}{c}{USPS $\rightarrow$ MNIST} \\\hline
Ours w/o incons &\textbf{93.2} & 89.4 & 91.7 & 82.1\\
Ours w/o label d & 79.3 & 95.4 & 95.1 & 92.9\\
Ours & 83.1 & \textbf{96.3} & \textbf{95.5} &\textbf{93.1}\\\hline
 &\multicolumn{4}{c}{MNIST $\rightarrow$ USPS} \\\hline
Ours w/o incons & 75.4 & 83.4 & 83.8 & 75.2\\
Ours w/o label d & 80.5 & 90.2 & 94.3 & \textbf{97.4}\\
Ours & \textbf{83.2} & \textbf{90.3} & \textbf{94.6} & 97.1\\
\bottomrule[1.5pt]
\end{tabular}}
\caption{Ablation studies for weighting with the target label distribution and the inconsistency loss. \textit{incons and label d} denote the inconsistency loss and label distribution based weighting respectively. $N$ denotes the number of classes in the target domain.}
\label{tb:ablation}
\end{center}
%\vspace{-1ex}
\end{table}

\noindent{\bf Ablation Study.}\ \ We investigate the effectiveness of weighting with the target label distribution and the inconsistency loss by using ablation. The first model involves the ablation of the inconsistency loss. The model is trained only with the weighted loss on the source.
The second model involves the ablation of target label distribution estimation and weighting. The model is trained with the inconsistency loss and the loss on the source without weighting with the target label distribution.
Tab.~\ref{tb:ablation} shows the results of adaptation between MNIST and USPS.  Although the accuracy of the model without label-distribution weighting drops at $N=3$, it performs well on the other setting. The results indicate that the inconsistency loss itself is effective for PDA. It is useful to combine the label distribution estimation when the number of target classes is small. 

\section{Conclusion}
In the study, we presented a novel method called Two Weighted Inconsistency-reduced Networks (TWINs) for partial domain adaptation (PDA). To align target samples with source classes present in the target domain, two classifiers estimate the label distribution in the target domain and weight classification loss. Furthermore, it learns discriminative features by minimizing inconsistency of two classifiers while inputting target samples. 
Our method outperformed existing methods with respect to several tasks in the PDA setting.

\section{Acknowledgement}
The work was partially supported by JST CREST Grant Number JPMJCR1403, Japan and was partially funded by the ImPACT Program of the Council for Science, Technology, and Innovation (Cabinet Office, Government of Japan). We
would like to thank Yusuke Mukuta, Yusuke Kurose, and Atsushi Kanehira for helpful
discussions.
%------------------------------------------------------------------------
%\clearpage
{\small
\bibliographystyle{ieee}
\bibliography{egbib}
}
\clearpage
\setcounter{section}{0}
\section*{Supplemental Material}
\setcounter{section}{0}

We would like to show supplementary information for our main paper. 
\section{Detail on experimental setting}
First, we introduce the detail of the experiments.
\subsection{Experiments on Digit and Traffic Sign Datasets}
\noindent{\textbf{Datasets.}}\ \ In all datasets, we use the standard training and test splits for training and testing respectively, and remove target samples of certain classes for the partial domain adaptation (PDA) setting.
MNIST~\cite{mnist} has 60000 training samples and 10000 test samples whose image size is  $28\times28$. USPS has 7291 training samples and 2007 test samples whose image size is $16\times16$. In the tasks of \textbf{MNIST $\to$ USPS} and \textbf{USPS $\to$ MNIST}, to match the image size, we rescale the USPS images to $28\times28$.
SVHN has 73257 training samples and 26032 test samples. In the task of \textbf{SVHN} $\to$ \textbf{MNIST}, we rescale MNIST images to $32\times32$. SYN SIGNS has 100000 samples and we randomly select 98000 samples for training samples. GTSRB has 39209 samples and we randomly select 31367 samples for training and the rest of 7842 samples for the test. Both of their images are $48\times48$ sizes.
\par
\noindent{\textbf{CNN architectures.}}\ \ In each domain adaptation tasks, we employ the three different CNN architectures used in~\cite{GRL}. We add Batch Normalization layers before the activation layers and dropout layers before the last free connected layer. In the dropout layers, a probability of an element to be zero is set as 0.5.\par
\noindent{\textbf{Optimization.}}
We use Adam optimizer~\cite{Adam} with a learning rate of $2.0 \times 10^{-4}$, a $\beta_1$ of 0.9, a $\beta_2$ of 0.999. Moreover, we set the coefficient of weight decay as $5\times10^{-4}$.
\par
\noindent{\textbf{Training procedure.}}\ \ As mentioned our main paper, we pre-train our classifiers by only source samples in the first 10 epochs (Phase~1), and we subsequently estimate the label distribution of target samples (Phase~2) and optimize the parameters of classifiers by source and target samples (Phase~3), repeatedly. Phase~2 is conducted after finishing Phase~1 and every time Phase~3 is repeated one epoch. 

\subsection{Experiments on \textbf{\textit{Office-31}} Datasets}
\noindent{\textbf{Datasets.}}\ \ As described in main paper, \textit{Office-31} has following three domains: \textit{Amazon}, \textit{Webcam}, and \textit{DSLR}. They have 2817, 795, and 488 samples, respectively.
%Following \cite{PADA}, after remove target samples belonging to certain classes for the PDA setting, we use all samples for training and report the accuracy on them.
\par
\noindent{\textbf{CNN architectures.}}\ \ We use PyTorch-provided models of ResNet-50~\cite{ResNet} pre-trained on ImageNet~\cite{ImageNet} as two classifiers following \cite{PADA}. The final fully connected layer is removed and replaced by a three-layered fully connected network. The dimension of the bottleneck layer is 256. To improve the performance of our method, we add dropout layer before the number of channels of CNN and between fully connected layers. In the dropout layers in CNN and fully connected layers, the dropout rate is set as 0.2 and 0.5, respectively.\par
\noindent{\textbf{Training procedure.}}\ \ As described the main paper, we do not use Phase~1 and begin training a model from Phase~3. We conduct Phase~2 every time we repeat Phase~3 500 iterations.

\begin{figure}[t] 
\centering
\subfigure[Source only]{\includegraphics[trim=0cm 3.5cm 0cm 3.5cm,clip,width=0.45\linewidth]{figure/source_emb.eps}
\label{fig:source_emb_}}
\subfigure[MCD]{\includegraphics[trim=0cm 3.5cm 0cm 3.5cm,clip,width=0.45\linewidth]{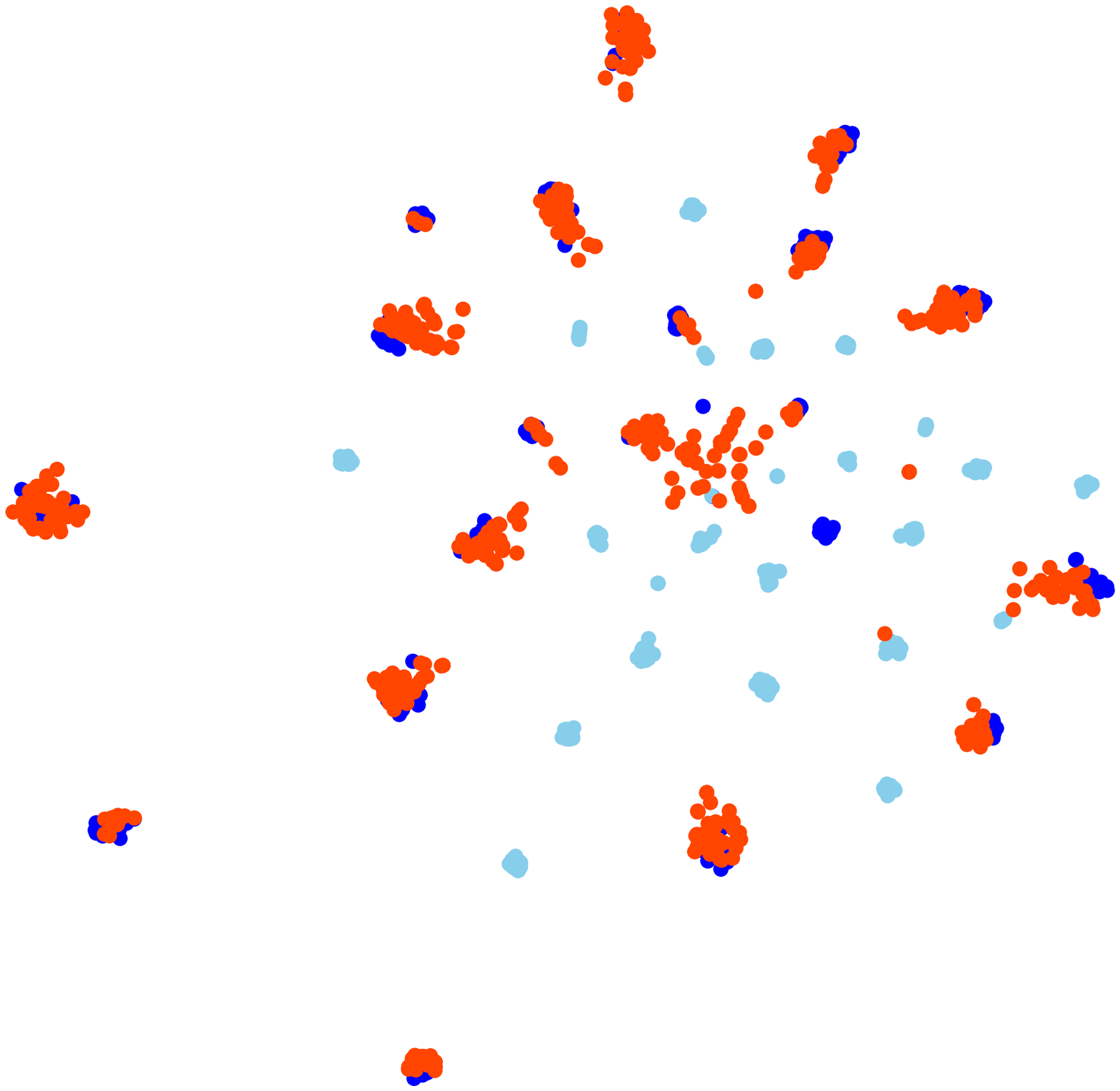}
\label{fig:MCD_emb_}}
\subfigure[PADA]{\includegraphics[trim=0cm 3.5cm 0cm 3.5cm,clip,width=0.45\linewidth]{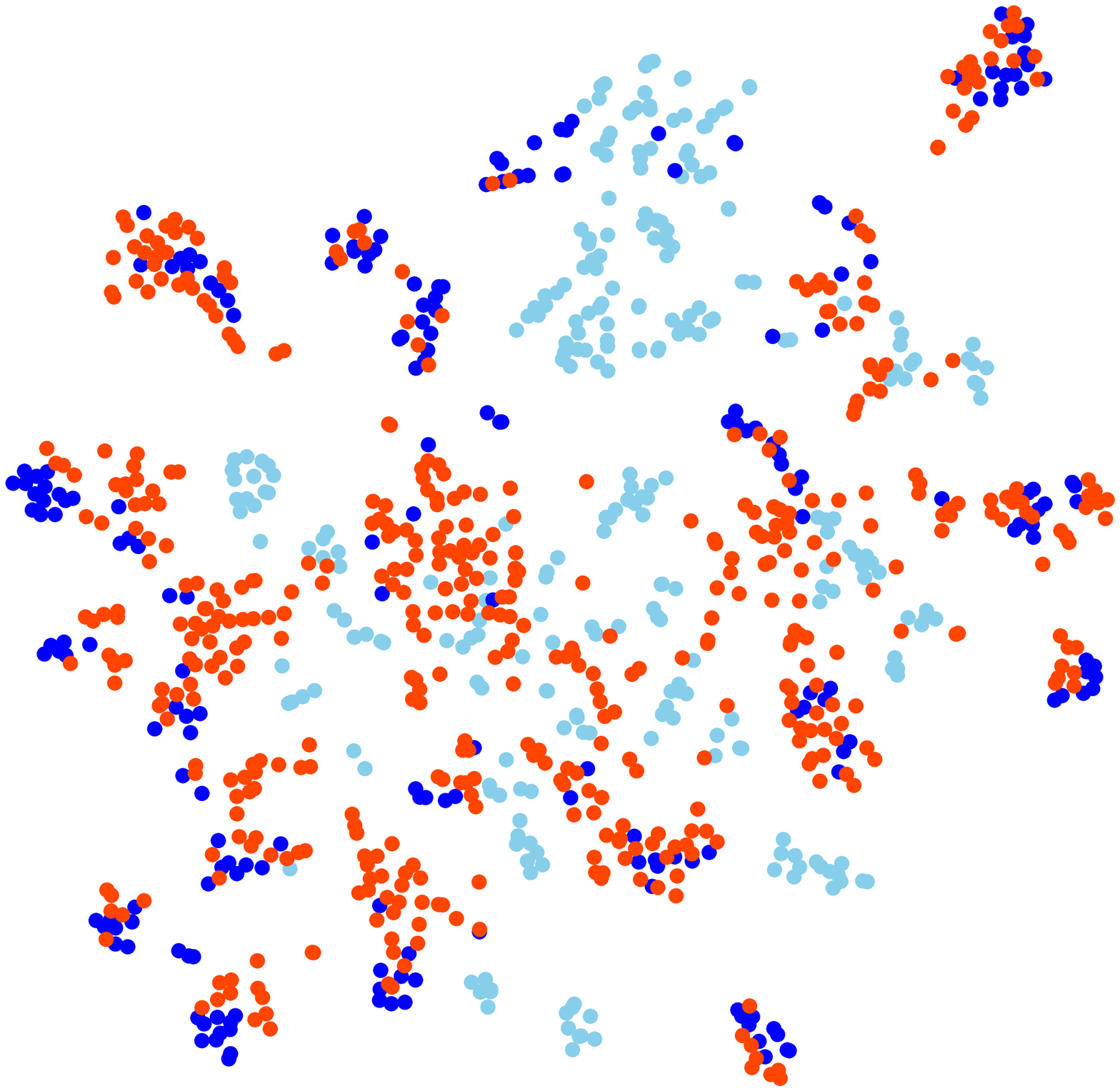}\label{fig:PADA_emb_}}
\subfigure[TWINs]{\includegraphics[trim=0cm 3.5cm 0cm 3.5cm,clip,width=0.45\linewidth]{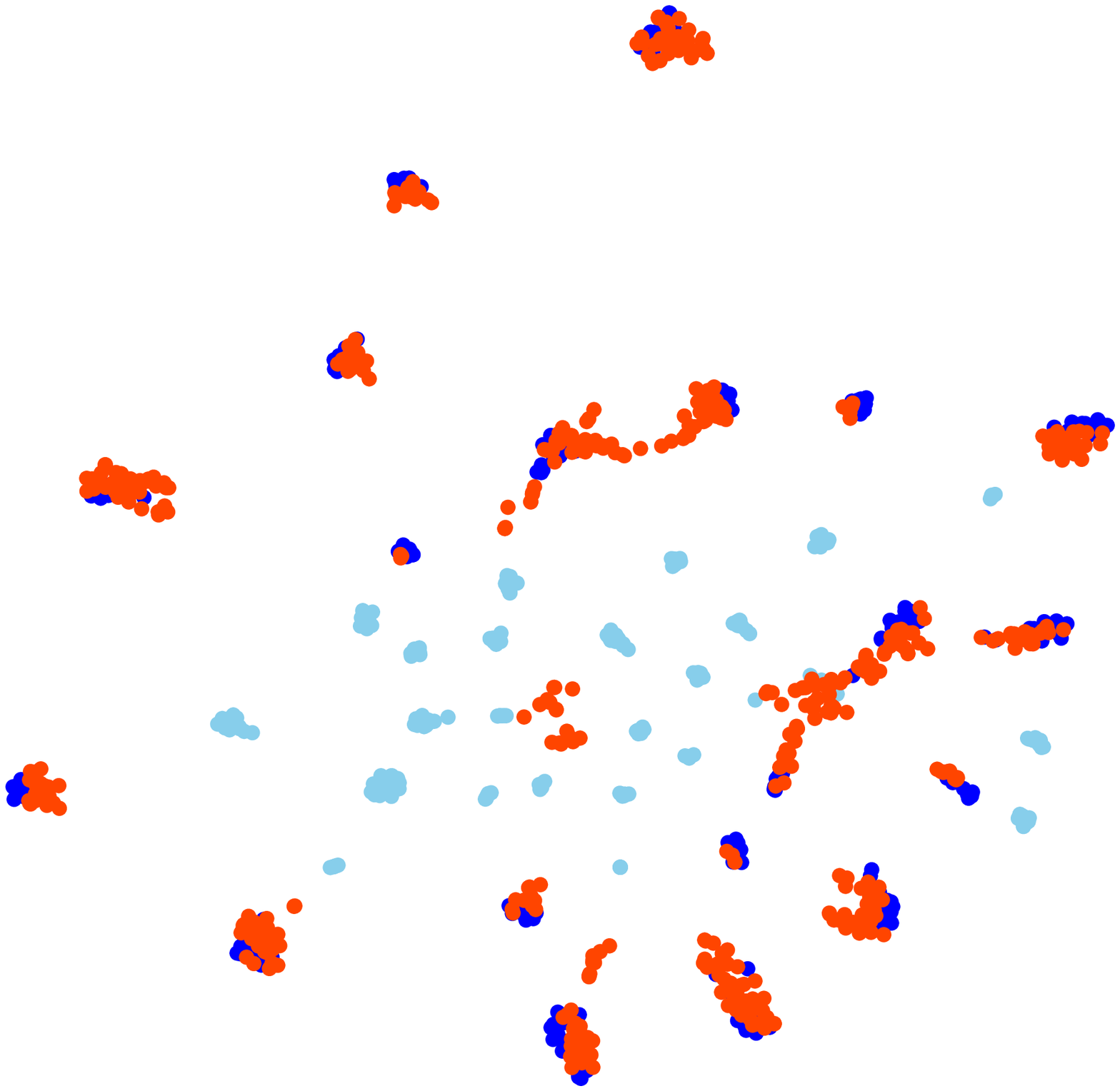}
\label{fig:twins_emb_}}
\caption{T-SNE visualization of features obtained from the second last fully connected layer of (a) source only, (b) MCD, (c) PADA, and (d) TWINs. The transfer task is SYN SIGNS (43 classes) $\to$ GTSRB (20 classes). Blue/light blue dots correspond to the source domain samples in which the classes are present/absent in the target domain while orange dots correspond to the target domain samples.}
\label{fig:t_SNE_}
\end{figure}

\section{Additional Analysis}
Finally, we show some additional analyses (\ie feature visualization and  learning curves) of our method.
\subsection{Feature Visualization}
We visualize feature distribution via t-SNE~\cite{t_SNE} in the different experiment from that showed in the main paper. The transfer task is SYN SIGNS (43 classes) $\to$ GTSRB (20 classes). Fig.~\ref{fig:t_SNE_} shows the visualized feature distribution. 
Fig.~\ref{fig:source_emb_}, Fig.~\ref{fig:MCD_emb_}, Fig.~\ref{fig:PADA_emb_}, and Fig.~\ref{fig:twins_emb_} show the feature distribution of source only, MCD~\cite{MCD}, PADA~\cite{PADA}, and TWINs, respectively. As discussed in the main paper, we can see the effectiveness of our proposed method thorough the visualization.

\begin{figure*}[t] 
\centering
\subfigure[SVHN $\to$ MNIST]{\includegraphics[clip,width=0.39\linewidth]{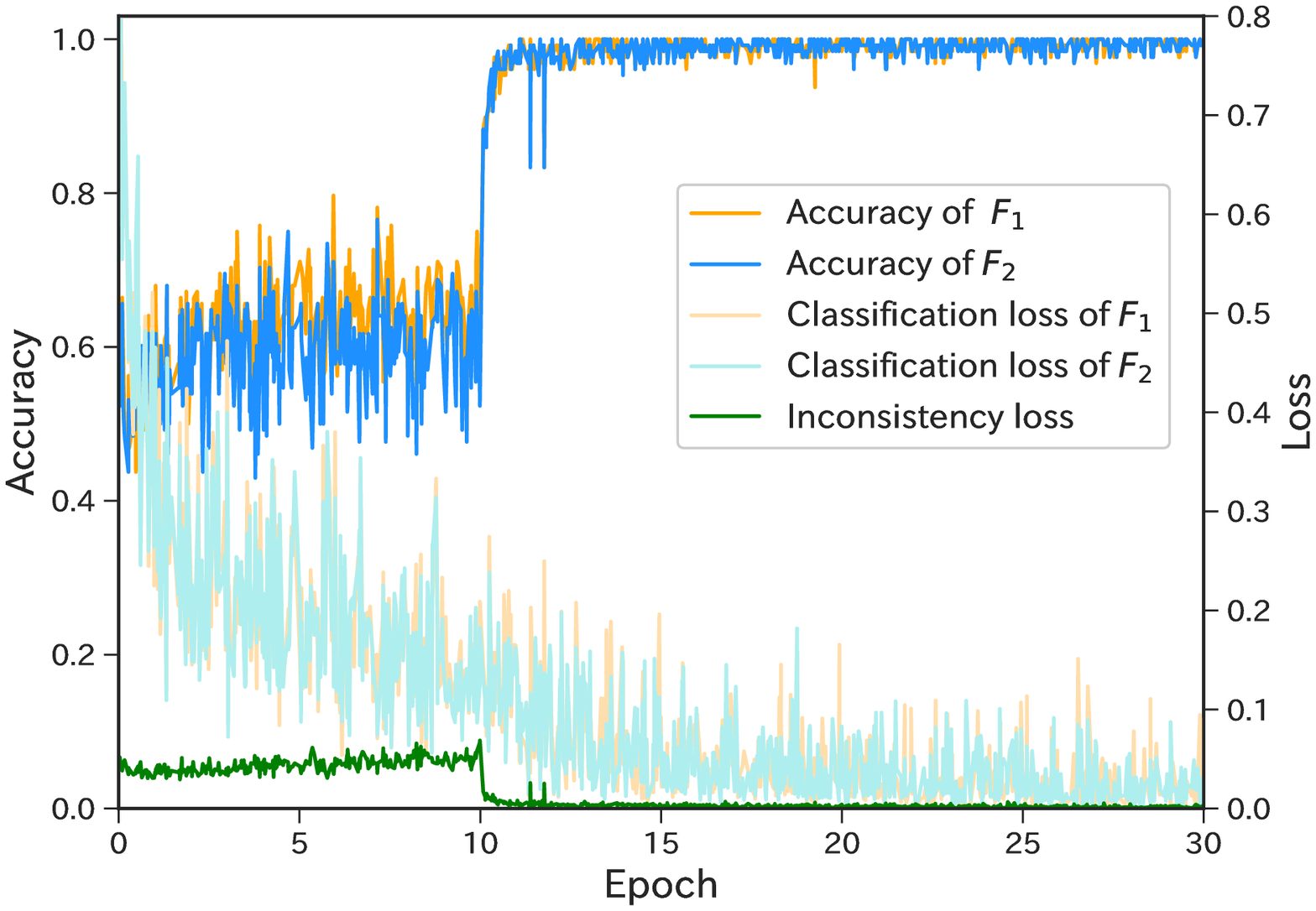}
\label{fig:loss_svhn2mnist}}
\subfigure[SYN SIGNS$\to$ GTSRB]{\includegraphics[clip,width=0.40\linewidth]{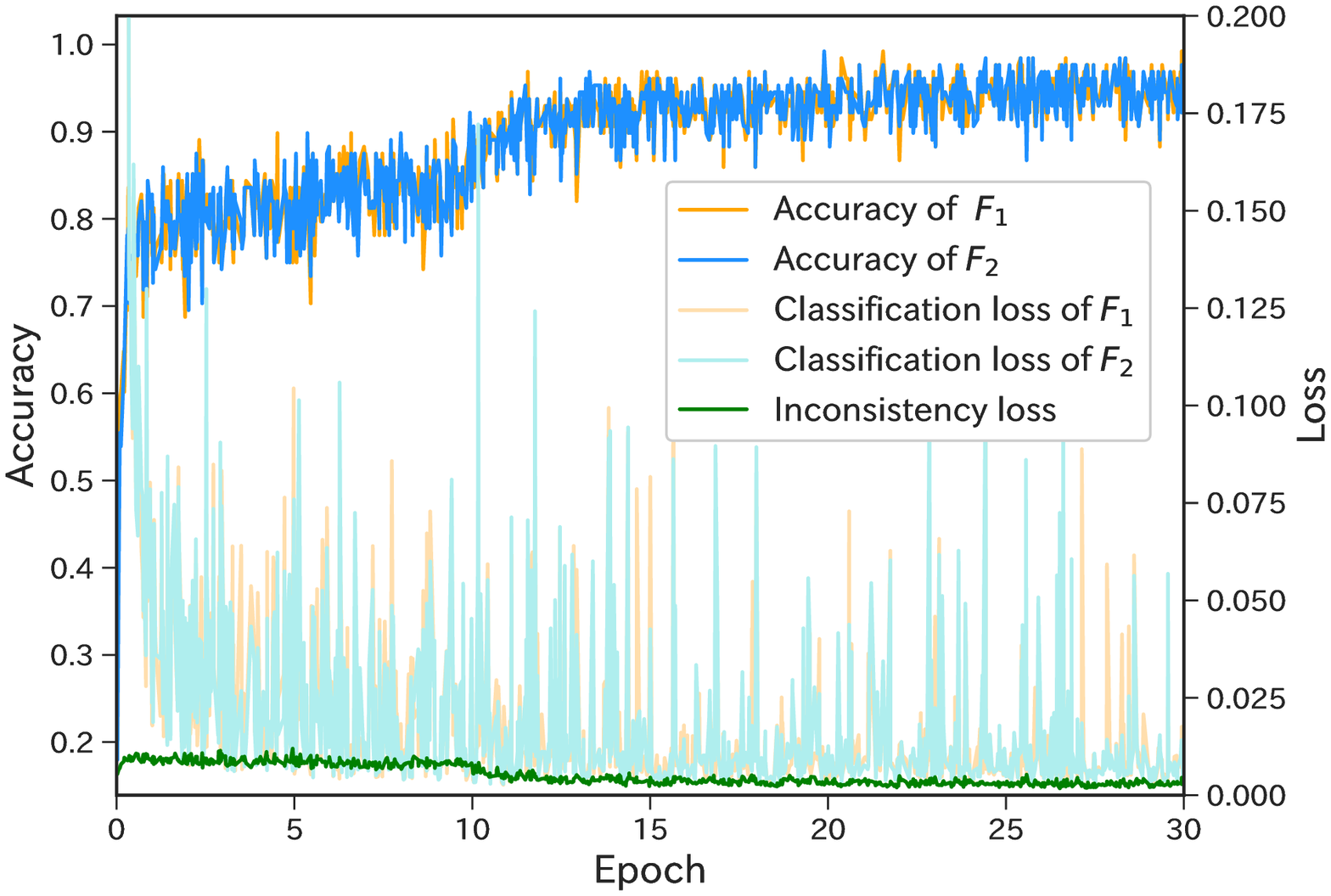}
\label{fig:loss_syn2gtsrb}}
\caption{Relationship among accuracies and losses during training. We pre-train $F_1,F_2$ by minimizing the classification losses on source samples for the first 10 epochs. After that, we minimize the weighted classification losses and the inconsistency loss.}
\label{fig:curve}
\end{figure*}

\subsection{Learning Curve}
We show the relationship among accuracies of the target samples and the loss during training in Fig.~\ref{fig:curve}. Fig.~\ref{fig:loss_svhn2mnist} and Fig.~\ref{fig:loss_syn2gtsrb} show the relationship on the adaptation from SVHN $\to$ MNIST and SYN SIGNS $\to$ GTSRB. We pre-train two classifiers $F_1$ and $F_2$ by minimizing the classification loss on source samples for the first 10 epochs (Phase~1). During this, the accuracies stay low due to the domain gap. After 10 epochs, we optimize the parameters of the classifiers by minimizing the weighted classification loss and the inconsistency loss (Phase~2 and Phase~3). Immediately after starting Phase~2 and Phase~3, the accuracies greatly improve, and the inconsistency loss decreases. 

\end{document}